\documentclass[11pt]{article}

\usepackage{acl}

\usepackage[utf8]{inputenc}
\usepackage[T1]{fontenc}
\usepackage{times}
\usepackage{latexsym}
\usepackage{hyperref}
\usepackage{url}
\usepackage{booktabs}
\usepackage{multirow}
\usepackage{amsfonts}
\usepackage{amsmath}
\usepackage{amssymb}
\usepackage{microtype}
\usepackage{xcolor}
\usepackage{graphicx}
\usepackage{subcaption}
\usepackage{float}
\usepackage{array}
\usepackage{tabularx}
\usepackage{placeins}
\usepackage{flushend}
\usepackage{dblfloatfix}
\usepackage{cuted}

\makeatletter
\apptocmd{\thebibliography}{\setlength{\itemsep}{-4.25pt plus 0.1pt minus 0.1pt}}{}{}
\makeatother

\newcommand{\Attn}{\mathrm{Attn}}       

\newcommand{\hz}[2]{\mathbf{z}^{#1}_{#2}}
\newcommand{\hl}[1]{\mathbf{z}^{#1}}

\newcommand{\Du}[2]{\Delta^{#1}_{#2}}
\newcommand{\Dl}[1]{\Delta^{#1}}
\newcommand{\Dup}[2]{\Delta^{#1}_{#2,\parallel}}
\newcommand{\Duperp}[2]{\Delta^{#1}_{#2,\perp}}

\newcommand{\pr}[2]{r^{#1}_{#2}}

\newcommand{\Adiag}[1]{A_{#1#1}}

\newcommand{\xsa}{XSA}

\newcommand{\spara}{s^{(\parallel)}}

\title{
Disentangling Representation Evolution in Transformers through Directional Decomposition}

\author{
  Shwai He$^{1,*}$ \quad
  Haichao Zhang$^{2}$ \quad
  Shen Yan$^{3}$ \\[2pt]
  $^{1}$University of Maryland, College Park \quad
  $^{2}$Northeastern University \quad
  $^{3}$ByteDance \\[2pt]
  \texttt{\small shwaihe@umd.edu} \quad
  \texttt{\small zhang.haich@northeastern.edu} \quad
  \texttt{\small sheny@bytedance.com}
}

\hypersetup{
  pdftitle={Disentangling Representation Evolution in Transformers through Directional Decomposition},
  pdfauthor={Shwai He, Haichao Zhang, and Shen Yan}
}

\begin{document}

\maketitle

\begingroup
\renewcommand{\thefootnote}{*}
\begin{NoHyper}
\footnotetext{Work done while Shwai He was an intern at ByteDance.}
\end{NoHyper}
\endgroup

\begin{abstract}
Transformer layers evolve representations through learned additive
transformations that combine parallel scaling and perpendicular steering.
Across pretrained models, we find that learned updates consistently contain
substantial parallel components. To test whether these components are
redundant or behaviorally important, we decompose updates in two complementary
spaces: residual space, analyzing complete sub-layer updates relative to the
incoming hidden state; and attention value space, analyzing
pre-output-projection aggregation relative to the token's own value. Targeted
edits reveal a consistent directional and space-dependent asymmetry:
perpendicular edits are consistently disruptive, whereas parallel edits are
comparatively redundant, especially in value space when preserving the direct
self message while scaling cross-token aggregation. The same geometry informs
compression and training: better-performing compressed models show lower
perpendicular transformation error, and suppressing parallel attention
components during from-scratch pretraining improves downstream performance across
evaluated model scales, with the value-space variant strongest. Overall, this geometry
connects update direction to editing robustness, compression diagnostics, and
training-time intervention. Code is available in the
\href{https://github.com/Shwai-He/Transformer-Geometry}{project repository}.
\end{abstract}

\section{Introduction}
\label{sec:intro}

Transformer layers evolve representations through learned additive
updates~\citep{vaswani2017attention,elhage2021mathematical} on top of residual
streams~\citep{he2016deep}. Geometrically, each update performs two distinct
roles relative to the incoming representation: a \textit{parallel component}
that modulates magnitude while preserving direction, and a \textit{perpendicular
component} that redirects it into new semantic subspaces. This dichotomy
reframes representation evolution as a problem of \textit{functional geometry}:
decomposing learned updates by their directional role and testing their
behavioral sensitivity through targeted interventions.

Across diverse pretrained models, our measurements reveal that learned updates
consistently contain substantial components parallel to their incoming states.
Geometrically, parallel updates amount to simple scalar rescaling, which the
residual stream already provides at zero parameter cost~\citep{he2016deep,elhage2021mathematical}.
Yet, heavily parameterized attention and MLP sub-layers actively allocate
capacity to produce such components. This paradox prompts a fundamental question:
are these parallel updates behaviorally essential to Transformer capabilities,
or are they functionally redundant?

To resolve this question, we formulate directional decomposition across two
complementary representation spaces: \textit{residual space}, which analyzes
sub-layer updates relative to the incoming hidden state, and \textit{attention
value space}, which analyzes internal value aggregation relative to the token's
own value representation. Using component-scaling interventions, we systematically
modulate the parallel and perpendicular components in these spaces to measure
their behavioral sensitivity.

Our interventions reveal a pronounced directional asymmetry across both
representation spaces. Perpendicular scaling is consistently disruptive,
confirming that orthogonal steering is essential to model capabilities.
In contrast, parallel scaling is comparatively benign, leaving performance
near baseline across a wide scaling interval. Within this parallel resilience,
the degree of stability depends on the representation space: in attention value
space, scaling the parallel component of cross-token aggregation keeps model
behavior remarkably close to baseline, whereas residual-space parallel scaling is
comparatively less stable. Thus, parallel updates exhibit broad resilience
compared to perpendicular steering, especially within internal value aggregation.

Finally, this directional geometry extends beyond inference-time editing to
post-training compression and from-scratch pretraining. For model
compression~\citep{Frantar2022GPTQAP,Lin2023AWQAW,Sun2023ASA,men2024shortgpt,He2026UncoveringTR},
decomposing update distortion reveals that better-performing quantized and
pruned models consistently exhibit lower perpendicular error, whereas parallel
error is far less discriminative. Motivated by this directional geometry, we
evaluate parallel attention suppression during pretraining. Across model
scales from 296M to 2.7B parameters, the value-space variant consistently
yields lower validation-loss trajectories, and both variants improve downstream
performance, with the value-space variant producing the largest gains.

\begingroup\setlength{\parskip}{0pt}In summary, the contributions of this work are as follows:
\begin{itemize}
  \setlength{\topsep}{0pt}
  \setlength{\partopsep}{0pt}
  \setlength{\itemsep}{2pt}
  \setlength{\parsep}{0pt}
  \setlength{\parskip}{0pt}
  \item This work reveals substantial direction-preserving components in learned
        transformer updates and maps their behavioral robustness across
        representation spaces.
  \item For attention updates, value-space decomposition separates direct
        self-value flow from cross-token aggregation, revealing a marked
        stability gap between value-space and residual-space edits.
  \item For compression and pretraining, this directional geometry extends beyond
        editing: perpendicular error tracks compressed-model quality, while
        parallel suppression consistently improves validation loss and
        downstream performance across model scales.
\end{itemize}
\endgroup

\section{Related Work}
\label{sec:related}

\paragraph{Residual Stream Magnitude Scaling.}
Residual connections preserve input representations while allowing each block
to add a learned transformation~\citep{he2016deep}. In transformers, this creates
a residual stream that carries information across layers and serves as the shared
workspace for attention and feed-forward updates~\citep{elhage2021mathematical}.
This preservation path is also a primary target for architectural scaling:
mechanisms such as ReZero~\citep{Bachlechner2020ReZeroIA}, LayerScale~\citep{touvron2021going},
and deep-transformer normalization schemes~\citep{xiong2020on,Wang2022DeepNetST}
adjust branch scale to stabilize optimization, while residual paths mitigate
representational degeneracy such as rank collapse~\citep{dong2021attention}.
These works establish the importance of residual pathways and branch magnitude
for training stability. In contrast, we study a complementary directional geometric property of the learned update itself: whether it reinforces the incoming representation
direction or steers it into orthogonal semantic subspaces. 

\paragraph{Self-Directed Attention Value Flow.}
Attention explicitly aggregates information across
tokens~\citep{vaswani2017attention}, but attention mass can also concentrate
on special or persistent positions, as shown by attention-sink behavior in
long-context inference~\citep{xiao2023efficient}. Gated attention variants
mitigate such sink behavior by modulating attention
outputs~\citep{qiu2025gatedattention}. However, even when attention sinks are reduced,
large diagonal attention weights can still route substantial mass back to the
current token itself. Exclusive Self-Attention (\xsa{}) directly addresses this
self-directed pathway by modifying value flow aligned with the current
token~\citep{zhai2025xsa}. In this work, we analyze this self-directed pathway from a directional geometric
perspective, disentangling direction-preserving updates from direction-changing
information.

\section{Background}
\label{sec:background}

\begin{figure*}[t]
  \centering
  \begin{subfigure}{0.49\textwidth}
    \centering
    \includegraphics[width=\linewidth]{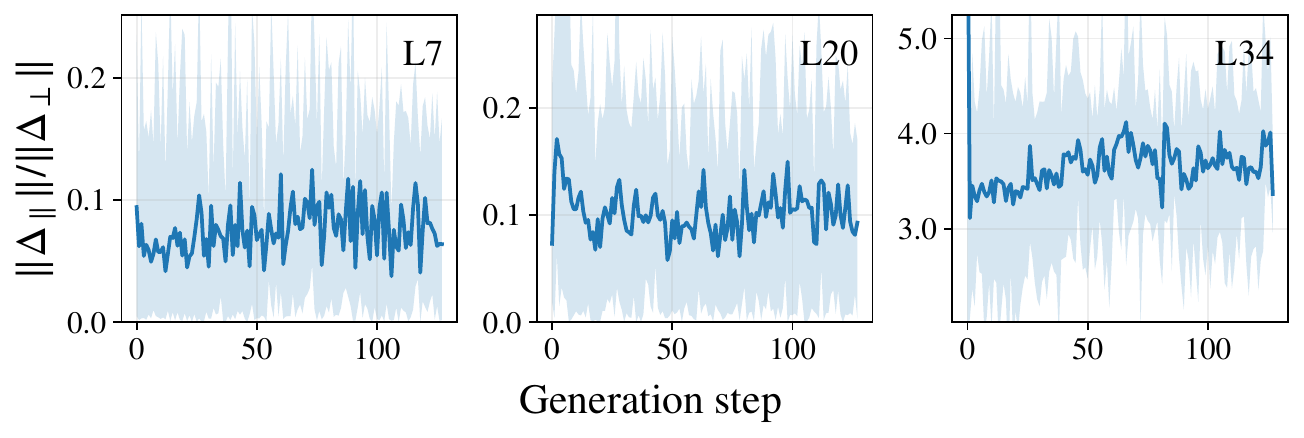}
    \caption{Qwen3-4B-Instruct-2507}
  \end{subfigure}\hfill
  \begin{subfigure}{0.49\textwidth}
    \centering
    \includegraphics[width=\linewidth]{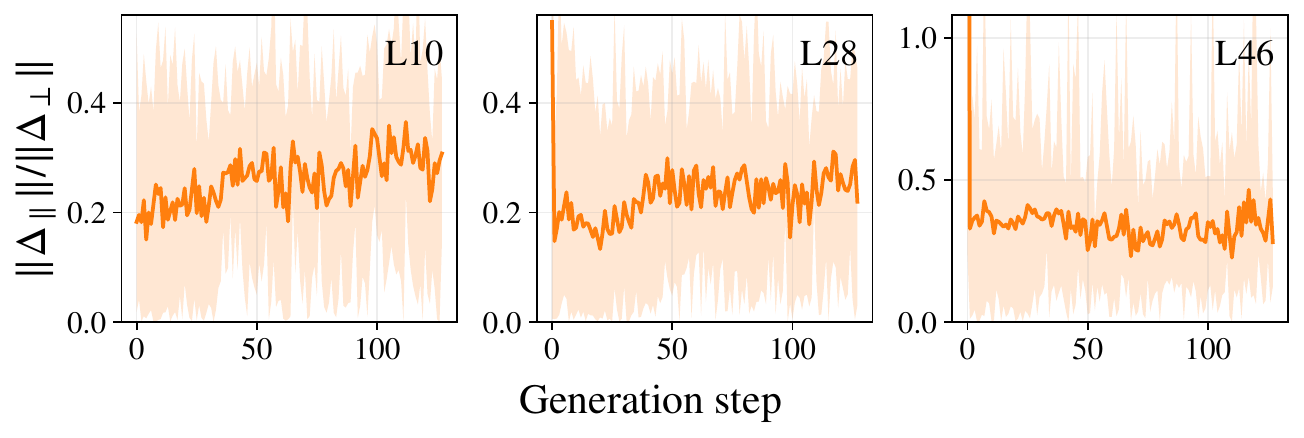}
    \caption{Qwen3-30B-A3B}
  \end{subfigure}
  \caption{\textbf{Parallel components persist across depth.}
  For each accumulated attention-plus-MLP update, $\Dup{l}{t}$ is its
  projection along the incoming hidden state, whereas $\Duperp{l}{t}$ is
  orthogonal to that state. Curves show the ratio
  $\|\Dup{l}{t}\|/\|\Duperp{l}{t}\|$ at three sampled layers per model
  across generation steps; lines are means and bands span the observed range.
  Ratios above one indicate parallel dominance.}
  \label{fig:profiles}
\end{figure*}

\subsection{Parallel and Perpendicular Decomposition}
\label{sec:decomp}

Given a reference vector $\mathbf{z} \in \mathbb{R}^d$ and an update vector
$\Delta \in \mathbb{R}^d$ within the same representation space, $\Delta$ admits
a unique orthogonal decomposition into components parallel and perpendicular to $\mathbf{z}$:
\begin{align}
  \Delta_\parallel
    &= \frac{\Delta \cdot \mathbf{z}}{\|\mathbf{z}\|^2}\,\mathbf{z}
     = \alpha\,\mathbf{z},
     \qquad
     \alpha = \frac{\Delta \cdot \mathbf{z}}{\|\mathbf{z}\|^2},
  \label{eq:para} \\
  \Delta_\perp
    &= \Delta - \Delta_\parallel,
    \qquad \Delta_\perp \perp \mathbf{z}.
  \label{eq:perp}
\end{align}
The parallel component $\Delta_\parallel$ lies along the reference direction
$\mathbf{z}$, scaling its magnitude by $\alpha$, whereas $\Delta_\perp$ lies in
the $(d-1)$-dimensional orthogonal subspace, carrying the direction-changing
degrees of freedom.

We apply this decomposition across two primary sites: in
\textit{residual space}, $\Delta$ is the sub-layer update added to the incoming
state $\mathbf{z}$; in \textit{attention value space}, $\Delta$ is the
pre-projection aggregate relative to the token's own value vector $\mathbf{z}$.

\subsection{Transformer Residual Updates}
\label{sec:transformer}

In a transformer with hidden dimension $d$, each residual sub-layer adds an update to the current representation:
\begin{equation}
  \hl{l+1} = \hl{l} + \Dl{l}.
  \label{eq:residual}
\end{equation}
The update $\Dl{l}$ denotes the output of an individual sub-layer
(self-attention $\Attn$ or $\mathrm{MLP}$) or the net update of the full
transformer block. The reference state $\hz{l}{t}$ is taken as the
representation immediately preceding the addition: the sub-layer input for
individual modules, or the block input for the accumulated update. We write
$\Du{l}{t}$ for the corresponding residual contribution at token position
$t \in \{1,\ldots,T\}$.

Using the decomposition in Equation~\ref{eq:para}, the parallel component of a residual update is:
\begin{equation}
  \Dup{l}{t} = \alpha^l_t \hz{l}{t}, \qquad
  \alpha^l_t = \frac{\Du{l}{t}\cdot \hz{l}{t}}{\|\hz{l}{t}\|^2},
\end{equation}
where the scalar $\alpha^l_t$ measures the magnitude of the parallel update relative to the incoming state $\hz{l}{t}$. Substituting this decomposition into the residual addition yields
\begin{equation}
  \hz{l+1}{t}
  = (1+\alpha^l_t)\hz{l}{t} + \Duperp{l}{t}, \qquad
  \Duperp{l}{t}\perp \hz{l}{t},
\end{equation}
where the parallel component rescales the incoming representation by $(1+\alpha^l_t)$, while $\Duperp{l}{t}$ introduces orthogonal direction-changing information.

Figure~\ref{fig:profiles} reports the relative magnitude $\|\Dup{l}{t}\|/\|\Duperp{l}{t}\|$ across generation steps at three sampled layers for two Qwen3 models. This ratio quantifies how much of a residual update aligns with the existing representation versus redirects it. The persistence of substantial parallel components across depth motivates investigating when they affect model behavior and in which representation spaces those effects remain robust.

\begin{figure*}[t]
  \centering
  \includegraphics[width=0.99\textwidth]{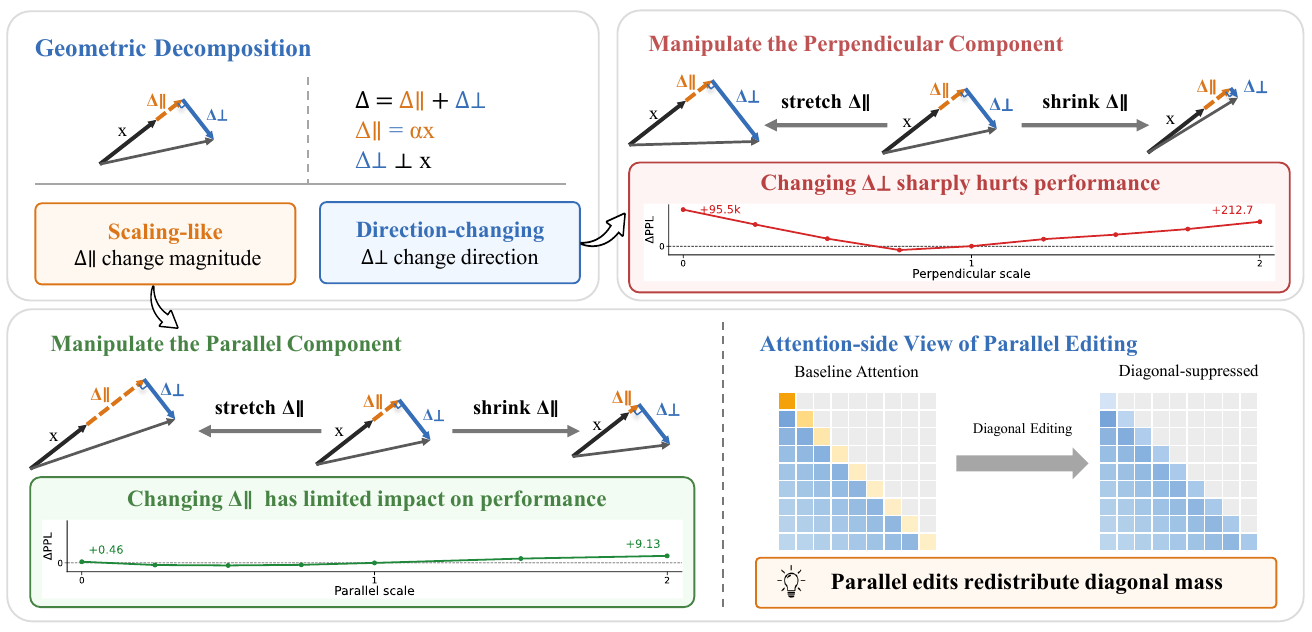}
  \caption{\textbf{Geometric decomposition and component scaling.}
  Residual updates and attention value aggregates are decomposed into parallel
  and perpendicular components and scaled independently; parallel-only scaling
  can be expressed as an attention-diagonal change. Labels report raw $\Delta\mathrm{PPL}$ relative to the
  scale-one no-op (lower is better); curves are rescaled only for display.}
  \label{fig:overview}
\end{figure*}

\section{Directional Interventions}
\label{sec:method}

We instantiate directional decomposition across two complementary spaces: the
residual stream (Section~\ref{sec:residual-level}) and the attention value space
(Section~\ref{sec:value-level}). We then unify these sites into a shared
component-scaling framework (Section~\ref{sec:closed-form-control}) to systematically
manipulate parallel and perpendicular updates. Figure~\ref{fig:overview}
summarizes both intervention sites and their diagonal view.

\subsection{Residual-Level Decomposition}
\label{sec:residual-level}

Set $\Delta = \Du{l}{t}$ (the residual update at layer $l$, token position $t$)
and $\mathbf{z} = \hz{l}{t}$ (the pre-update hidden state).
For layer and module profiles, we report calibration-corpus averages of the
per-token residual parallel ratio:
\begin{equation}
  \begin{aligned}
  \pr{l}{t}
    &= r\!\left(\Du{l}{t},\,\hz{l}{t}\right)
     = \frac{|\alpha^l_t|\,\|\hz{l}{t}\|}{\|\Duperp{l}{t}\|},\\
  \alpha^l_t
    &= \frac{\Du{l}{t}\!\cdot\!\hz{l}{t}}{\|\hz{l}{t}\|^2}.
  \end{aligned}
  \label{eq:para-ratio}
\end{equation}
This residual-level view serves as the shared observable for the rest of the
paper. The resulting diagnostic is block-agnostic. We apply it to
(i)~the $\Attn$ sub-layer output,
(ii)~the $\mathrm{MLP}$ sub-layer output, and
(iii)~their accumulated effect at the transformer-block level.

\subsection{Value-Space Decomposition}
\label{sec:value-level}

Inside attention, let $\mathbf{x}^{\mathrm{attn}}_s$ denote the input to the
value projection, including the model's pre-attention normalization. Each
source token $s$ is mapped to a value vector
$\mathbf{v}_s=W_V\mathbf{x}^{\mathrm{attn}}_s$. For query token $t$, let
$\mathcal{A}_{ts}$ denote a linear attention-mixing operator on the value
space. This notation does not assume a particular head structure; standard
multi-head attention is the block-diagonal instance whose head blocks carry
their corresponding attention weights.
The pre-output-projection aggregate is
\begin{equation}
  \mathbf{o}_t=\sum_{s\le t} \mathcal{A}_{ts}\mathbf{v}_s .
  \label{eq:value-aggregation}
\end{equation}
Within this space the natural reference is the current token's own value
$\mathbf{v}_t$: the direction $\mathbf{o}_t$ would take if attention routed
nothing from other positions.
\begin{equation}
  \mathbf{o}_{t,\parallel}
    = \frac{\mathbf{o}_t \cdot \mathbf{v}_t}{\|\mathbf{v}_t\|^2}\mathbf{v}_t,
  \qquad
  \mathbf{o}_{t,\perp}=\mathbf{o}_t-\mathbf{o}_{t,\parallel}.
  \label{eq:value-space-decomp}
\end{equation}
Residual-space geometry does not isolate this self-value-aligned structure
because the output projection mixes the aggregate into the residual stream.

\paragraph{Direct Self-Message Preservation.}
In naive value-space decomposition (Equation~\ref{eq:value-space-decomp}), the
query token's self message $\mathbf{d}_t = \mathcal{A}_{tt}\mathbf{v}_t$ is
colinear with $\mathbf{v}_t$ and absorbed into $\mathbf{o}_{t,\parallel}$.
Naive parallel removal ($s^{(\parallel)}=0$) thus extinguishes the token's
identity carrier alongside cross-token features. To isolate \textit{contextual}
magnitude modulation while preserving self-representation, we decouple the aggregate:
\begin{equation}
  \begin{aligned}
  \mathbf{o}_t
    &= \underbrace{\mathcal{A}_{tt}\mathbf{v}_t}_{\mathbf{d}_t\;\text{(self)}}
     + \underbrace{\sum_{s<t}\mathcal{A}_{ts}\mathbf{v}_s}_{\mathbf{c}_t\;\text{(non-self)}},\\
  \mathbf{c}_{t,\parallel}
    &= \frac{\mathbf{c}_t \cdot \mathbf{v}_t}{\|\mathbf{v}_t\|^2}\mathbf{v}_t,
  \qquad
  \mathbf{c}_{t,\perp}=\mathbf{c}_t-\mathbf{c}_{t,\parallel}.
  \end{aligned}
  \label{eq:exclude-self-decomp}
\end{equation}
We scale only the parallel and perpendicular components of the non-self
aggregate and then restore the unchanged self message:
\begin{equation}
  \widetilde{\mathbf{o}}^{\mathrm{excl}}_t
    = \mathbf{d}_t
      +s^{(\parallel)}\mathbf{c}_{t,\parallel}
      +s^{(\perp)}\mathbf{c}_{t,\perp}.
  \label{eq:exclude-self-scaling}
\end{equation}
We refer to this operation as \emph{exclude-self} scaling: it neither masks the
self-attention edge nor renormalizes the attention row.

Value-space decomposition applies only to attention aggregates. For MLPs, the
main analysis decomposes the sub-layer output in residual space; Appendix~\ref{app:mlp-internal-carrier}
separately applies the same projection geometry to the post-gating carrier and
grouped down-projected contributions.

\subsection{Component Scaling and Diagonal Form}
\label{sec:closed-form-control}

For either attention-side site, write
$\mathbf y_t=\sum_{s\le t}\mathcal M_{ts}\mathbf m_s$ and decompose
$\mathbf y_t$ relative to its reference $\mathbf r_t$. Value-space editing
uses $(\mathcal M_{ts},\mathbf m_s,\mathbf r_t)
=(\mathcal A_{ts},\mathbf v_s,\mathbf v_t)$. Residual-space attention editing
uses $\mathcal B_{ts}=W_O\mathcal A_{ts}$ and
$(\mathcal M_{ts},\mathbf m_s,\mathbf r_t)
=(\mathcal B_{ts},\mathbf v_s,\hz{l}{t})$, keeping head-dependent mixing
explicit. Component scaling then uses the shared intervention
\begin{equation}
  \widetilde{\mathbf{y}}_t
    =s^{(\parallel)}\mathbf{y}_{t,\parallel}
     +s^{(\perp)}\mathbf{y}_{t,\perp}.
  \label{eq:unified-component-scaling}
\end{equation}
The no-op is $s^{(\parallel)}=s^{(\perp)}=1$, and parallel removal sets
$s^{(\parallel)}=0,s^{(\perp)}=1$. Exclude-self scaling applies the same form
to $\mathbf{c}_t$ and restores $\mathbf{d}_t$ as in
Equation~\ref{eq:exclude-self-scaling}.

The same edit can also be read on the attention map itself, connecting it to
self-directed attention mass (Section~\ref{sec:related}). Holding all off-diagonal
messages fixed, a diagonal update realizes parallel-only scaling when
\begin{equation}
  \Delta\mathbf y_{tt}
    =\bigl(s^{(\parallel)}-1\bigr)\mathbf y_{t,\parallel}.
  \label{eq:general-diagonal-constraint}
\end{equation}
Appendix~\ref{app:method} gives an operator realization, scalar
attention-diagonal forms, removed cases, and the applied-output audit.

\begin{figure*}[t]
  \centering
  \includegraphics[width=0.98\linewidth]{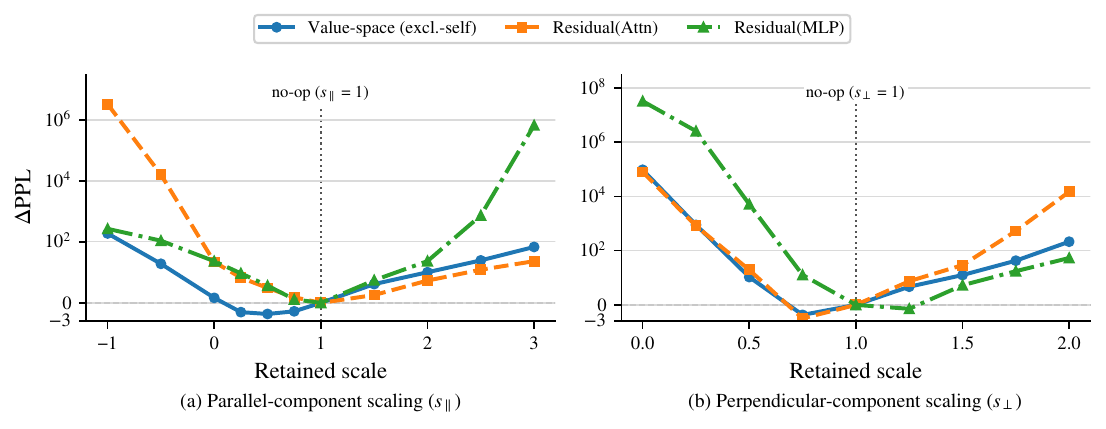}
  \caption{\textbf{Perplexity under component scaling.}
  The left panel varies $s_{\parallel}$ with $s_{\perp}=1$; the right varies
  $s_{\perp}$ with $s_{\parallel}=1$. Evaluated on Qwen3-1.7B (WikiText-2),
  $\Delta\mathrm{PPL}$ is edited minus the matched scale-one no-op, so lower is
  better. All decoder layers are edited; value-space curves use exclude-self
  scaling.}
  \label{fig:residual-component-editing}
\end{figure*}

\section{Experimental Setup}
\label{sec:experiments}
\label{sec:setup}

\paragraph{Intervention implementation.}
All inference-time edits operate strictly during the forward pass and leave
model weights untouched. We apply component scaling across decoder layers in
two distinct sites: the residual stream and attention value space.
In value space, we explicitly evaluate both \textit{full-aggregate} scaling
(decomposing the complete attention aggregate $\mathbf{o}_t$ relative to $\mathbf{v}_t$)
and \textit{exclude-self} scaling (Equation~\ref{eq:exclude-self-scaling}).
Because the query token's direct self message $\mathbf{d}_t = \Adiag{t}\mathbf{v}_t$
is intrinsically parallel to $\mathbf{v}_t$, naive parallel removal on the full
aggregate inadvertently suppresses the token's primary identity carrier along with
cross-token updates, causing severe degradation (Section~\ref{sec:value-vs-residual-exp}).
Exclude-self scaling preserves $\mathbf{d}_t$ to isolate whether cross-token contextual
magnitude modulation is truly redundant. Unless explicitly marked otherwise,
value-space evaluations use exclude-self scaling.
Appendices~\ref{app:evaluation-details} and~\ref{app:method} provide complete
evaluation and implementation details.

\paragraph{Models and benchmarks.}
We evaluate across dense and mixture-of-experts architectures from the
Qwen3~\citep{Yang2025Qwen3TR}, Llama-3~\citep{Dubey2024TheL3}, and
Gemma-3~\citep{GemmaTeam2025Gemma3} families.
Our benchmarks include language modeling perplexity, seven standard zero-shot
commonsense and reasoning tasks, long-context retrieval and reasoning (13-task
RULER suite~\citep{Hsieh2024RULERWT} up to 12k context lengths), and
post-training model compression under 4-bit AWQ~\citep{Lin2023AWQAW} and Wanda
pruning~\citep{Sun2023ASA} (unstructured, 4:8, and 2:4).

\paragraph{Pretraining setup.}
For training-time investigations, we pretrain GPT-style models from scratch on
OpenWebText~\citep{Gokaslan2019OpenWeb} across scales from 296M to 2.7B parameters,
comparing baseline optimization against residual- and value-space
parallel removal; Appendix~\ref{app:training-configs} details architectural
configurations and training hyperparameters.

\section{Inference-Time Component Editing}
\label{sec:editing-diagnostics}

\subsection{Directional Component Scaling}
\label{sec:zeroshot}

\paragraph{Probing directional sensitivity.}
Section~\ref{sec:method} formalized the orthogonal decomposition into parallel
magnitude modulation and perpendicular semantic steering across residual and
value spaces. We first evaluate the functional roles of these components by
measuring model sensitivity to continuous inference-time scaling edits.
Holding weights frozen, Figure~\ref{fig:residual-component-editing} independently
varies $s^{(\parallel)}$ (left, $s^{(\perp)}=1$) and $s^{(\perp)}$ (right,
$s^{(\parallel)}=1$) across three intervention sites:
value space (exclude-self), residual attention, and residual MLP.

\paragraph{Directional asymmetry: acute perpendicular fragility.}
Across all three intervention sites, model behavior exhibits an acute
\textit{directional asymmetry}. Perpendicular scaling ($s^{(\perp)} \ne 1$)
proves extraordinarily hyper-fragile: even minor deviations from identity
immediately destabilize the model with steep perplexity surges, while severe
attenuation or complete removal ($s^{(\perp)}=0$) triggers catastrophic
breakdown, escalating by multiple orders of magnitude across all sites (reaching
tens of thousands of points in value space and residual attention, and exploding
into millions of points in residual MLP). This confirms that orthogonal
directions govern delicate, non-interchangeable semantic steering. In stark
contrast, parallel scaling ($s^{(\parallel)}$) displays a broad, stable tolerance
basin, confirming parallel updates regulate representation magnitude
rather than categorical semantic trajectories.

\paragraph{Site asymmetry: value space vs.\ residual branches.}
Within the parallel axis, the sweeps reveal a pronounced \textit{site
hierarchy}. Value-space manipulation (with the direct self message preserved)
stays closest to baseline, remaining virtually flat across
$s^{(\parallel)} \in [0, 1]$ (shifting PPL by less than a point, and by at most
1.3 points up to scale 3). In the residual stream, Residual(Attn) is
intermediate, whereas Residual(MLP) is most fragile, collapsing rapidly
outside the unit interval.

This hierarchy demonstrates that robustness is not an inherent property of
``parallelness'' in the abstract, but depends decisively on representation space
and what is preserved. Residual-space parallel edits modify the accumulated
features of the main stream, where parallel updates carry essential cross-layer
computation. In contrast, value-space exclude-self editing modulates only
cross-token contextual aggregation while leaving the query token's primary
identity carrier ($\mathbf{d}_t = \Adiag{t}\mathbf{v}_t$) and perpendicular context
intact. A parallel effect appears inside the MLP: decomposing the internal
post-gating carrier rather than the final sub-layer output similarly preserves
performance under parallel removal (Appendix~\ref{app:mlp-internal-carrier}).

\subsection{Cross-Space Representation Editing}
\label{sec:value-vs-residual-exp}

\paragraph{Robustness on general tasks.}
We evaluate downstream zero-shot robustness across seven standard benchmarks on
dense (Qwen3-1.7B) and MoE (Qwen3-30B-A3B) models (Table~\ref{tab:general-zeroshot}),
comparing three primary approaches: (1) self-preserving value-space removal
(V-Excl.-self), (2) naive full-aggregate value removal (V-Para Rem.), and (3)
residual-space and diagonal controls (Attn Para-Rem.\ and Diag.\ Rem.).

When parallel removal is applied naively to the entire value aggregate (V-Para Rem.),
average performance drops by 8--10 points across both models. This degradation
illustrates why isolating self-attention is essential: naive value-space editing
suppresses the token's own identity carrier $\Adiag{t}\mathbf{v}_t$ alongside
cross-token messages, inadvertently disrupting representation propagation.

In stark contrast, preserving the direct self-message while scaling only
cross-token aggregation (V-Excl.-self) leaves performance virtually intact across
all seven benchmarks—trailing baseline by only 1.5 points on 1.7B and virtually
matching it on the 30B-A3B MoE model (within 0.1 points), with slight gains on
several individual tasks. Meanwhile, residual-space attention removal (Attn
Para-Rem.) incurs steady drops across tasks, and hard diagonal removal (Diag.\ Rem.)
severely degrades the models because forcing $\Adiag{t}=0$ artificially renormalizes
attention weights over off-diagonal positions. These comparisons
establish that cross-token parallel updates are functionally redundant, provided
direct self-representations are retained.

\begin{table*}[t]
  \caption{\textbf{Task-level comparison of residual-, value-space, and diagonal controls.}
  Entries are absolute scores (\%); higher is better and Avg. is unweighted.
  Both value-space rows set $s_{\parallel}=0$. V-Para Rem. edits the full
  value aggregate; V-Excl.-self preserves $\Adiag{t}\mathbf{v}_t$. Diag.\ Rem.\
  zeros $\Adiag{t}$ and renormalizes over non-self positions.
  Attn Para-Rem.\ and Diag.\ Rem.\ use separate matched
  baselines. Differences reported in the text are computed before rounding.
  Bold marks the best baseline/value-space result.}
  \label{tab:general-zeroshot}
  \centering
  \scriptsize
  \setlength{\tabcolsep}{3.1pt}
  \resizebox{\textwidth}{!}{%
  \begin{tabular}{llrrrrrrrr}
    \toprule
    Model & Setting & OBQA & PIQA & Wino & ARC-C & HSwag & BoolQ & RTE & \underline{Avg.} \\
    \midrule
    \multirow{5}{*}{Qwen3-1.7B}
      & Baseline & 37.20 & \textbf{72.47} & \textbf{60.69} & \textbf{53.07} & \textbf{60.16} & \textbf{77.65} & \textbf{70.04} & \underline{\textbf{61.61}} \\
      & Attn Para-Rem. & 38.20 & 72.31 & 56.59 & 50.09 & 58.74 & 71.38 & 66.79 & \underline{59.16} \\
      & V-Para Rem. & 34.40 & 69.48 & 54.85 & 41.38 & 54.04 & 66.27 & 53.43 & \underline{53.41} \\
      & V-Excl.-self & \textbf{39.20} & 71.76 & 60.54 & 49.15 & 59.88 & 74.46 & 65.70 & \underline{60.10} \\
      & Diag. Rem. & 26.20 & 51.36 & 49.17 & 24.49 & 26.66 & 40.70 & 55.60 & \underline{39.17} \\
    \midrule
    \multirow{5}{*}{Qwen3-30B-A3B}
      & Baseline & \textbf{44.80} & \textbf{80.69} & 70.24 & \textbf{69.97} & 77.79 & \textbf{88.53} & 81.95 & \underline{\textbf{73.42}} \\
      & Attn Para-Rem. & 44.00 & 78.73 & 66.85 & 65.10 & 73.87 & 88.04 & 78.34 & \underline{70.70} \\
      & V-Para Rem. & 41.20 & 76.22 & 64.56 & 59.04 & 69.57 & 70.89 & 62.45 & \underline{63.42} \\
      & V-Excl.-self & 44.60 & 79.87 & \textbf{72.45} & 68.34 & \textbf{78.13} & 86.36 & \textbf{83.39} & \underline{73.31} \\
      & Diag. Rem. & 30.60 & 57.67 & 48.46 & 26.96 & 34.78 & 38.59 & 47.65 & \underline{40.68} \\
    \bottomrule
  \end{tabular}
  }
\end{table*}

\paragraph{Context-length scalability.}
Table~\ref{tab:longcontext-ruler} reinforces this distinction as context expands
(RULER on Llama-3.2-3B from 4k to 12k). Retaining half of the parallel component
($\spara=0.5$) with the self-message fixed closely tracks the unedited baseline
across all sequence lengths (within 0.2 points at 4k and 2.4 points at 12k),
consistently outperforming full-aggregate scaling. Under complete removal
($\spara=0$), full-aggregate editing collapses by 26--40 points, whereas
exclude-self retains over $75\%$ accuracy at 4k and maintains substantially higher
resilience out to 12k. Table~\ref{tab:appendix-ruler-full} details 13-task breakdowns and multi-model evaluations.

\begin{table}[!htbp]
  \caption{\textbf{Long-context performance across sequence lengths.}
  Reported scores are mean accuracy (\%) across all 13 RULER tasks on
  Llama-3.2-3B for context lengths from 4k to 12k tokens. V-Full scales
  the entire value aggregate; V-Excl.-self preserves direct self messages
  $\mathbf{d}_t$ and scales only cross-token context.}
  \label{tab:longcontext-ruler}
  \centering
  \footnotesize
  \renewcommand{\arraystretch}{0.95}
  \begin{tabular*}{\linewidth}{@{\extracolsep{\fill}}lcrrr}
    \toprule
    Method & $s_\parallel$ & 4k & 8k & 12k \\
    \midrule
    Baseline & $1.0$ & 86.91 & 82.09 & 79.30 \\
    V-Full & $0.5$ & 84.81 & 78.80 & 74.76 \\
    V-Excl.-self & $0.5$ & \textbf{86.69} & \textbf{80.80} & \textbf{76.88} \\
    V-Full & $0.0$ & 60.99 & 48.61 & 39.17 \\
    V-Excl.-self & $0.0$ & 75.64 & 67.97 & 62.02 \\
    \bottomrule
  \end{tabular*}
\end{table}

\begin{figure}[tb]
  \centering
  \includegraphics[width=0.97\linewidth]{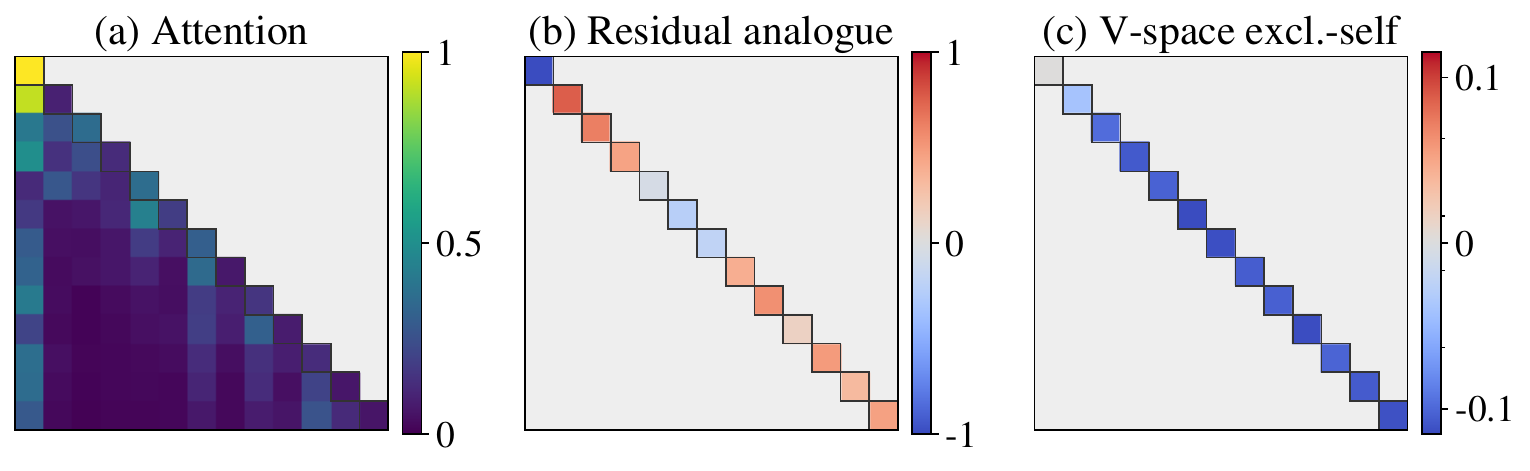}
  \caption{\textbf{Attention diagonal views of parallel editing.}
  For full parallel removal ($s_{\parallel}=0$), the left panel shows empirical
  attention weights (Qwen3-4B, layer 19, head 6); the middle shows the residual-space
  attention-map view, and the right shows the exclude-self value-space change.
  The latter two color scales encode $\Delta A_{tt}$; off-diagonal entries are unchanged.}
  \label{fig:attn-maps}
\end{figure}

\paragraph{Attention diagonal views for parallel editing.}
\label{sec:diagonal-validation}
All three attention-side interventions admit closed-form expressions as effective
diagonal modifications on the attention matrix (Section~\ref{sec:closed-form-control}).
Figure~\ref{fig:attn-maps} compares the empirical attention map of Qwen3-4B
(layer 19, head 6) under baseline against the residual-space and value-space
diagonal views (see Table~\ref{tab:effective-diagonal-derivation} in the Appendix
for complete mathematical derivations).

The residual-space diagonal view induces volatile positive and negative adjustments
(spanning $-1.00$ to $0.76$), whereas exclude-self value-space editing preserves
direct self-token routing with bounded, non-positive diagonal shifts.
Crucially, this localized stability does not mean value-space editing is gentler:
auditing the post-$W_O$ attention output on Qwen3-0.6B reveals
that removing parallel context across non-self tokens induces a much larger
perturbation norm than residual-space removal ($49.10\%$ versus $25.15\%$ of
unedited branch norm; Table~\ref{tab:applied-no-para-strength}).
That value-space editing preserves model capabilities despite altering attention
outputs nearly twice as much proves that stability is governed by preserving
self-identity routing and residual propagation trajectories, rather than minimizing
isotropic perturbation magnitude.
Appendix~\ref{app:head-localization} details head-level localization.

\subsection{Compression Diagnostics}
\label{sec:compression-error-geometry}

\begin{figure*}[t]
  \centering
  \includegraphics[width=0.98\textwidth]{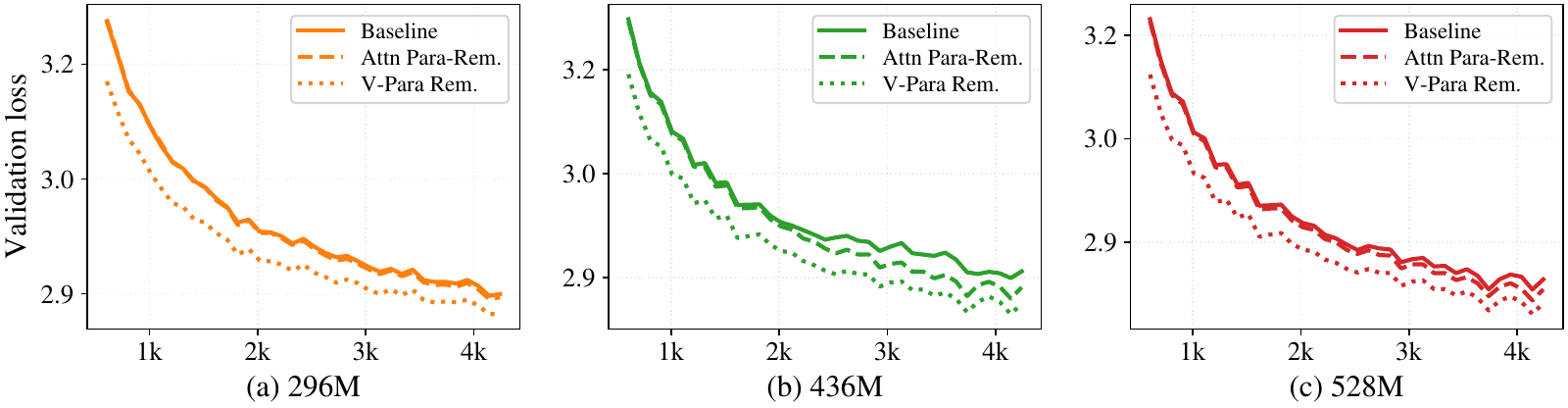}
  \caption{\textbf{Training-time parallel removal lowers validation loss.}
  The displayed OpenWebText~\citep{Gokaslan2019OpenWeb} loss trajectories
  show baseline, Attn Para-Rem., and V-Para Rem. for matched configurations across
  296M, 436M, and 528M model sizes. Both interventions set $\spara=0$ and retain
  the perpendicular component.}
  \label{fig:pretraining}
  \vspace{-0.8em}
\end{figure*}

The removal experiments establish that model capabilities are exceptionally
sensitive to direction-changing components of activation updates. We next ask
how this geometric principle manifests under post-training compression. Compression
alters model parameters, inducing an error between uncompressed and compressed
updates rather than an explicit activation edit. Decomposing this error into
parallel and perpendicular components provides a principled test: if orthogonal
steering is the behaviorally critical component, superior compression algorithms
should systematically exhibit lower perpendicular distortion, whereas parallel error
should be far less diagnostic.

\begin{figure}[t]
  \centering
  \includegraphics[width=0.98\linewidth]{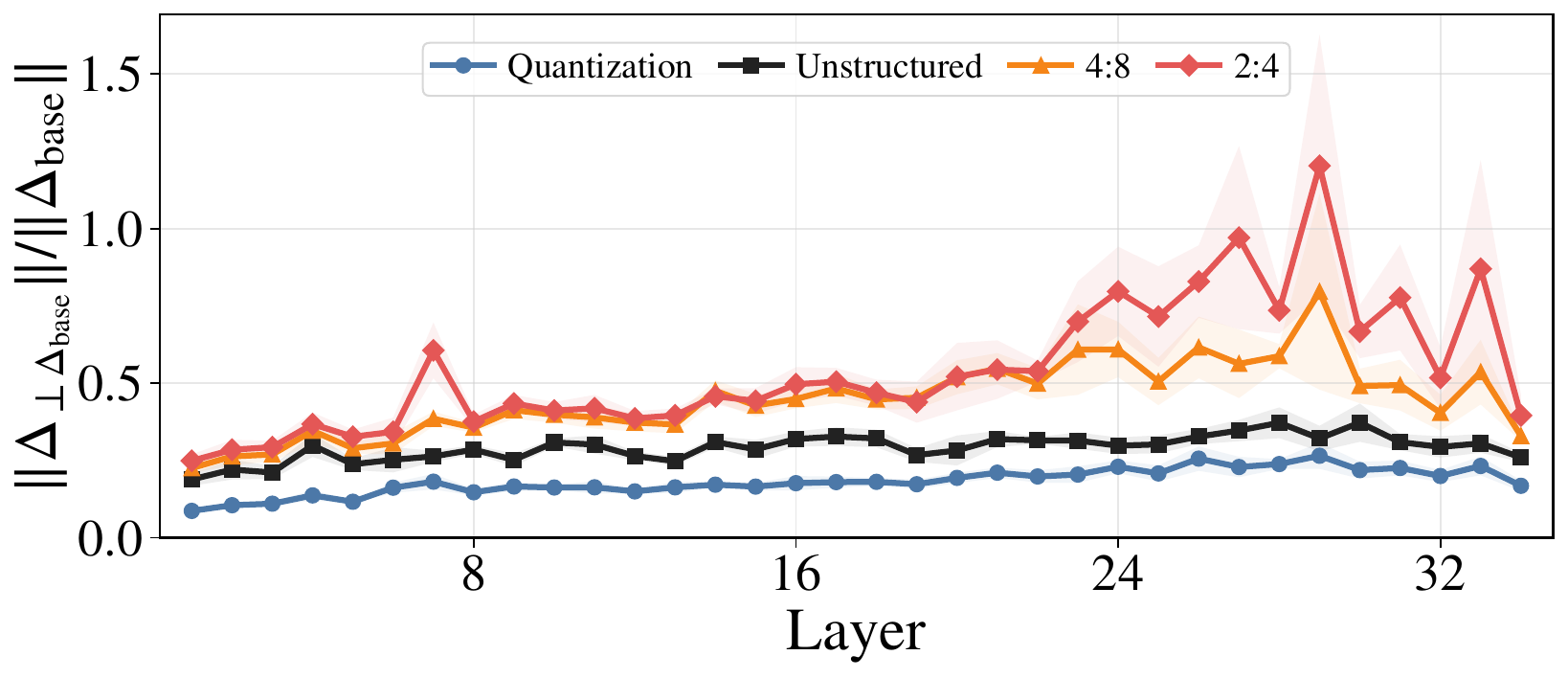}\par
  {\small (a) Perpendicular attention-output error.\par}

  \vspace{0em}
  \includegraphics[width=0.98\linewidth]{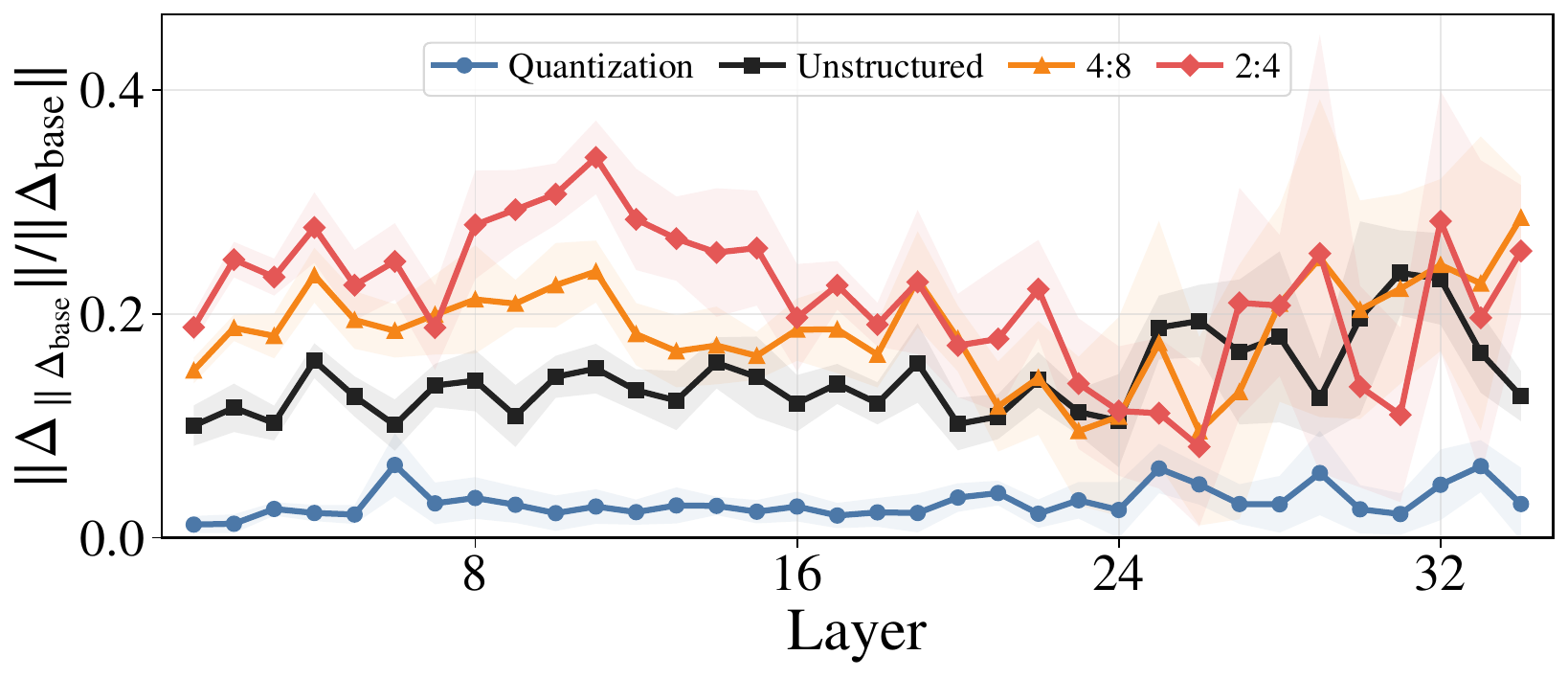}\par
  {\small (b) Parallel attention-output error.\par}
  \caption{\textbf{Directional decomposition of compression error.}
  Curves are layerwise means, with bands showing one standard deviation over 12 fixed prompts,
  comparing compressed and baseline Qwen3-4B attention updates. The upper and
  lower panels show perpendicular and parallel error, each normalized by the
  full baseline-update norm (higher is more distortion). Settings are
  4-bit AWQ and 50\% Wanda pruning with unstructured, 4:8, or 2:4 masks.}
  \label{fig:local-flip-attn}
  \vspace{-1em}
\end{figure}

Concretely, at each layer, let $\Delta_{\mathrm{base}}$ denote the uncompressed
baseline update and $\Delta_{\mathrm{comp}}$ denote the compressed update. The local
compression error is $e = \Delta_{\mathrm{comp}} - \Delta_{\mathrm{base}}$. We
decompose $e$ relative to the baseline update direction:
$e_{\parallel} = \mathrm{proj}_{\Delta_{\mathrm{base}}} e$ and
$e_{\perp} = e - e_{\parallel}$. Figure~\ref{fig:local-flip-attn} reports the
normalized component norms $\|e_{\perp}\|/\|\Delta_{\mathrm{base}}\|$ and
$\|e_{\parallel}\|/\|\Delta_{\mathrm{base}}\|$ across layers of Qwen3-4B under
4-bit AWQ and 50\% Wanda pruning (unstructured, 4:8, and 2:4).

The results strongly corroborate this directional hypothesis: perpendicular error
curves cleanly separate the four compression regimes in strict alignment with their
downstream fidelity (4-bit AWQ achieves the lowest error, followed by unstructured,
4:8, and 2:4 Wanda). In contrast, parallel error curves heavily interleave and fail
to provide a consistent ranking. This clarifies why conventional isotropic $L_2$ error
can fail to predict degradation: isotropic distance conflates benign magnitude shifts
with disruptive semantic rotations. Appendix~\ref{app:compression} substantiates
this mechanism across the MLP branch and full block updates, showing that perpendicular
distortion dominates total compression error ($r > 0.97$, Table~\ref{tab:local-total-perp-alignment})
and connecting this geometry to cosine-based pruning diagnostics.

\section{Training-Time Allocation}
\label{sec:pretraining}

In standard Transformer pretraining, attention updates conflate colinear magnitude
rescaling with orthogonal contextual steering. \textit{We investigate whether suppressing
parallel attention updates throughout training shifts how optimization allocates
representational capacity across direction-changing components.} We train GPT-style
language models from scratch on standard next-token prediction, enforcing parallel
removal during both optimization and evaluation to eliminate train--evaluation mismatch.
We evaluate these interventions through validation trajectories and downstream
benchmarks; Appendix~\ref{app:training-configs} gives model configurations and
gated-scaling controls.

Figure~\ref{fig:pretraining} plots OpenWebText~\citep{Gokaslan2019OpenWeb} validation loss
trajectories across model sizes (296M, 436M, and 528M). Suppressing parallel
attention updates shifts the learning trajectory downward from early training
stages, and this advantage persists consistently through the final checkpoint
rather than emerging only as an endpoint artifact.

Table~\ref{tab:pretraining-summary} reports downstream zero-shot accuracy across
six standard benchmarks for the scaled 1.4B and 2.7B models. Both interventions
improve downstream performance over the baseline at both scales, with value-space
parallel removal achieving the largest gains ($+0.7$ points on 1.4B and $+1.5$ points on 2.7B).
Together with training trajectories, these findings demonstrate that
directional decomposition provides an effective inductive bias: suppressing
parallel updates directs attention capacity toward orthogonal contextual steering,
improving downstream generalization.

\begin{table}[t]
  \caption{\textbf{Downstream generalization across model scales.}
  Unweighted mean accuracy over six benchmarks (ARC-Easy, BoolQ, HellaSwag,
  OpenBookQA, PIQA, WinoGrande); \(\Delta\)Avg is relative to matched baseline.}
  \label{tab:pretraining-summary}
  \centering
  \small
  \renewcommand{\arraystretch}{0.96}
  \begin{tabular*}{0.92\columnwidth}{@{\extracolsep{\fill}}llrr}
    \toprule
    Model & Setting & Avg & $\Delta$Avg \\
    \midrule
    \multirow{3}{*}{1.4B}
      & Baseline & 58.5 & 0.0 \\
      & Attn Para-Rem. & 58.8 & +0.3 \\
      & V-Para Rem. & \textbf{59.2} & \textbf{+0.7} \\
    \midrule
    \multirow{3}{*}{2.7B}
      & Baseline & 60.2 & 0.0 \\
      & Attn Para-Rem. & 60.9 & +0.7 \\
      & V-Para Rem. & \textbf{61.7} & \textbf{+1.5} \\
    \bottomrule
  \end{tabular*}
\end{table}

Appendix~\ref{app:training-configs} details task-level scores for both larger
models and the gated-scaling control, where fixed removal maintains the lowest
training loss.

\section{Discussion}
\label{sec:discussion}

\paragraph{Geometric Duality of Representation Evolution.}
Our decomposition reveals that Transformer layer updates execute two distinct geometric functions: \textit{magnitude modulation} ($\Delta_\parallel$) and \textit{semantic steering} ($\Delta_\perp$). Perpendicular updates rotate representations into orthogonal subspaces housing factual and syntactic distinctions; because these coordinates are finely calibrated, $\Delta_\perp$ is acutely fragile. Conversely, parallel updates act as adaptive gain controllers, amplifying existing trajectories without altering directional meaning. The extensive tolerance basin of $\Delta_\parallel$ confirms that Transformers possess inherent resilience to magnitude variations, preserving semantic intent despite substantial scale disruption.

\paragraph{Beyond Isotropic Compression Error.}
Quantization and pruning minimize isotropic $L_2$ error $\min \|\Delta - \widehat{\Delta}\|_2^2$, assuming uniform sensitivity across $\mathbb{R}^d$. Our directional framework exposes this flaw: update-aligned errors ($\Delta_\parallel^{\mathrm{err}}$) fall within the benign tolerance basin, whereas orthogonal deviations ($\Delta_\perp^{\mathrm{err}}$) directly distort semantic trajectories. This explains why compression methods with comparable global $L_2$ error diverge in downstream fidelity. Measuring directional distortion establishes $\Delta_\perp^{\mathrm{err}}$ as a causal predictor of degradation, showing compression objectives should penalize directional rotation over uniform distance.

\paragraph{Architectural Implications and Pretraining Dynamics.}
In standard self-attention, token mixing is structurally entangled with self-representation amplification, allocating parameter capacity to redundant parallel scaling. Suppressing parallel updates relieves attention from these duties, guiding optimization to dedicate capacity toward directional contextual steering. This inductive bias accelerates convergence, lowering validation-loss trajectories and improving downstream generalization across evaluated model scales. These findings provide principled geometric support for emerging architectures that decouple contextual routing from magnitude modulation (such as gated attention and query-key normalization).

\section{Conclusion}
\label{sec:conclusion}

We formalize representation evolution in Transformers through orthogonal decomposition, separating updates into parallel magnitude modulation ($\Delta_\parallel$) and perpendicular semantic steering ($\Delta_\perp$). Across diverse pretrained language models, interventions reveal a universal directional asymmetry: representations tolerate substantial colinear scaling along their trajectory, yet exhibit acute fragility under perpendicular perturbations. Furthermore, this stability is strongly site-dependent: value-space aggregation proves markedly more robust than residual-space edits by preserving direct token routing.

This geometric perspective resolves key challenges across the model lifecycle. In compression, directional error decomposition clarifies why isotropic $L_2$ distance often fails to predict behavioral loss: perpendicular distortion causally drives degradation, whereas parallel shifts are largely benign. In pretraining, suppressing parallel attention updates acts as an effective inductive bias, directing capacity toward orthogonal contextual routing. Across evaluated model scales, this constraint lowers validation loss and improves downstream performance. By connecting directional geometry to representation dynamics, our findings offer the NLP community a principled framework to move beyond isotropic Euclidean heuristics in model diagnosis, compression, and architecture design.

\section*{Limitations}
\label{sec:limitations}

Our experiments cover decoder-only language models across multiple scales,
language tasks, inference-time interventions, compression settings, and
from-scratch training configurations. Although the same geometric picture is
consistent across the evaluated settings, its behavior under other
architectures and training regimes remains to be established.
Future work should test whether these directional patterns persist across such
settings and connect layer-level geometry more directly to downstream behavior.
We use the proposed edits primarily as analysis tools; turning them into
practical training or inference methods requires further validation.

\bibliography{references}

\appendix
\clearpage
\raggedcolsend
\raggedend
\raggedbottom

\setlength{\textfloatsep}{6pt plus 2pt minus 1pt}
\setlength{\floatsep}{4pt plus 2pt minus 1pt}
\setlength{\intextsep}{7pt plus 2pt minus 1pt}
\setlength{\dbltextfloatsep}{7pt plus 2pt minus 1pt}
\setlength{\dblfloatsep}{5pt plus 2pt minus 1pt}
\makeatletter
\setlength{\@dblfptop}{0pt}
\setlength{\@dblfpsep}{10pt}
\setlength{\@dblfpbot}{0pt plus 1fil}
\makeatother

\section{Evaluation Details}
\label{app:evaluation-details}

\paragraph{Evaluated models.}
The editing and diagnostic experiments use Qwen3~\citep{Yang2025Qwen3TR} base
checkpoints, except Figure~\ref{fig:profiles}, which uses Qwen3-4B-Instruct-2507.
Long-context evaluation uses Llama-3.2-3B~\citep{Dubey2024TheL3}.

\paragraph{Benchmark evaluation.}
Tasks are evaluated via Language Model Evaluation Harness~\citep{eval-harness}
and RULER~\citep{Hsieh2024RULERWT}, following standard configurations~\citep{Wang2018GLUEAM,
Hendrycks2020MeasuringMM,Cobbe2021TrainingVT,reddy-etal-2019-coqa,dua-etal-2019-drop,
lin-etal-2022-truthfulqa,Clark2018ThinkYH,Zellers2019HellaSwagCA}.
Table~\ref{tab:evaluation-task-settings} gives the default few-shot setting and
metric for each task.

\begin{table}[H]
  \centering
  \scriptsize
  \setlength{\tabcolsep}{1.8pt}
  \caption{\textbf{Default evaluation task settings.} Acc. denotes accuracy;
  Norm denotes length normalization.}
  \label{tab:evaluation-task-settings}
  \begin{tabular}{@{}lcl@{\hspace{6pt}}lcl@{}}
    \toprule
    Task & Shots & Metric & Task & Shots & Metric \\
    \midrule
    ARC-Chal. & 25 & Acc. (Norm) & MMLU & 5 & Acc. \\
    ARC-Easy & 0 & Acc. (Norm) & OpenBookQA & 0 & Acc. (Norm) \\
    BoolQ & 0 & Acc. & PIQA & 0 & Acc. (Norm) \\
    CoQA & 0 & F1 & RTE & 0 & Acc. \\
    DROP & 0 & F1 & TruthfulQA-MC & 0 & Acc. \\
    GSM8K & 5 & Exact match & TruthfulQA gen. & 0 & Rouge-L \\
    HellaSwag & 10 & Acc. (Norm) & WinoGrande & 5 & Acc. \\
    \bottomrule
  \end{tabular}
  \normalsize
\end{table}

\section{Intervention Details}
\label{app:method}

\paragraph{Attention-diagonal editing.}
Equation~\ref{eq:general-diagonal-constraint} defines the requested change in the parallel component.
Let $\mathbf m_t$ be the self message, $\mathbf r_t$ the reference, and $\mathcal B_{tt}$ its linear map into the edited space.
One minimum-Frobenius-norm realization of $\Delta\mathbf y_{tt}=\Delta\mathcal B_{tt}\mathbf m_t$ is
\(\Delta\mathcal{B}_{tt}^{\star} = \bigl(s^{(\parallel)}-1\bigr)
\frac{\mathbf{y}_t^\top\mathbf{r}_t}{\|\mathbf{m}_t\|^2\|\mathbf{r}_t\|^2}\mathbf{r}_t\mathbf{m}_t^\top\).
Projecting onto the admissible operator subspace yields the scalar forms in Table~\ref{tab:effective-diagonal-derivation}.
For our primary exclude-self value-space edit, Eq.~\ref{eq:exclude-self-scaling} applies projection to \(\sum_{s<t}A_{ts}\mathbf{v}_s\), restoring direct \(A_{tt}\mathbf{v}_t\); isolating self-interaction from contextual mixing suppresses redundant cross-token parallel drift while preserving the diagonal identity carrier, ensuring downstream stability.

\paragraph{Applied strength of the two edits.}
Section~\ref{sec:diagonal-validation} and Table~\ref{tab:effective-diagonal-derivation}
show that exclude-self value-space editing preserves the self contribution
$A_{tt}\mathbf{v}_t$, inducing bounded non-positive shifts along the attention diagonal,
whereas residual-space removal forces volatile signed swings ($-1.00$ to $+0.76$).
However, diagonal stability does not imply a gentler intervention.
Table~\ref{tab:applied-no-para-strength} audits both Full No-Para interventions at the
post-$W_O$ output, comparing edited $\widetilde{\mathbf{y}}_t$ against baseline $\mathbf{y}_t$.
The exclude-self value-space edit induces a substantially larger perturbation
on every metric: $49.10\%$ norm change and $25.89\%$ energy removal, versus
$25.15\%$ and $9.32\%$ for residual-space removal.
This divergence reveals a geometric asymmetry: residual-space removal
prunes only the 1D projection along incoming $\hz{}{t}$, which is a smaller fraction of branch
norm yet fatally disrupts inter-layer residual propagation.
Conversely, value-space editing zeros parallel components across the non-self
aggregate $\sum_{s<t} A_{ts}\mathbf{v}_s$, discarding significant contextual energy
while preserving the primary diagonal identity channel.
Preserving capabilities despite perturbing outputs nearly twice as heavily
shows that stability depends on structural alignment rather than isotropic magnitude.

\begin{table}[!t]
  \vspace{-6pt}
  \caption{\textbf{Attention-diagonal forms for component scaling.}
  Value-space form is exact; residual form matches scalar constraint;
  ``removed case'' sets $s^{(\parallel)}=0$ while retaining perpendicular component.}
  \label{tab:effective-diagonal-derivation}
  \centering
  \scriptsize
  \setlength{\tabcolsep}{2pt}
  \begingroup
  \renewcommand{\arraystretch}{0.86}
  \begin{tabularx}{\linewidth}{>{\raggedright\arraybackslash}p{0.28\linewidth}X}
    \toprule
    Quantity & Form \\
    \midrule
    \multicolumn{2}{c}{\textit{Full-aggregate value space}} \\
    \cmidrule(lr){1-2}
    Diagonal change &
    \(\Delta A_{tt}=(s^{(\parallel)}-1)
      \frac{\mathbf{o}_t^\top\mathbf{v}_t}{\|\mathbf{v}_t\|^2}\). \\
    Removed case &
    \(\widetilde A_{tt}=
      -\sum_{s<t}A_{ts}
      \frac{\mathbf{v}_s^\top\mathbf{v}_t}{\|\mathbf{v}_t\|^2}\). \\
    \midrule
    \multicolumn{2}{c}{\textit{Residual space}} \\
    \cmidrule(lr){1-2}
    Diagonal change &
    \(\Delta A_{tt}=(s^{(\parallel)}-1)
      \frac{(\Delta_t^{\Attn})^\top\hz{}{t}}
           {(W_O\mathbf{v}_t)^\top\hz{}{t}}\). \\
    Removed case &
    \(\widetilde A_{tt}=A_{tt}
      -\sum_{s\le t}A_{ts}
      \frac{(W_O\mathbf{v}_s)^\top\hz{}{t}}
           {(W_O\mathbf{v}_t)^\top\hz{}{t}}\). \\
    \bottomrule
  \end{tabularx}
  \endgroup
  \vspace{-6pt}
\end{table}

\begin{table}[H]
  \vspace{-2pt}
  \centering
  \scriptsize
  \setlength{\tabcolsep}{3pt}
  \caption{\textbf{Applied strength of attention Full No-Para interventions.}
  Full No-Para sets $s_{\parallel}=0$ on Qwen3-0.6B. Perturbation norm is
  $\|\widetilde{\mathbf{y}}-\mathbf{y}\|/\|\mathbf{y}\|$, retained norm is
  $\|\widetilde{\mathbf{y}}\|/\|\mathbf{y}\|$, and perturbation energy is
  $\|\widetilde{\mathbf{y}}-\mathbf{y}\|^2/\|\mathbf{y}\|^2$ relative to baseline $\mathbf{y}$.}
  \label{tab:applied-no-para-strength}
  \begingroup
  \renewcommand{\arraystretch}{0.88}
  \resizebox{\columnwidth}{!}{%
  \begin{tabular}{lrrr}
    \toprule
    Geometry & Perturb.\ norm & Perturb.\ energy & Retained norm \\
    \midrule
    Attention residual-space & 25.15\% & 9.323\% & 95.01\% \\
    Attention exclude-self value-space & 49.10\% & 25.89\% & 83.36\% \\
    \bottomrule
  \end{tabular}}
  \endgroup
  \vspace{-2pt}
\end{table}

\paragraph{Computational cost.}
We apply Eq.~\ref{eq:unified-component-scaling} via forward hooks replacing only the target tensor.
A fused FlashAttention-2~\citep{dao2024flashattention2} kernel combines projections and attention.
As Table~\ref{tab:projection-overhead} reports, overhead is negligible ($<2.5\%$ in prefill, $<1.0\%$ in decoding, $3.39\%$ in pretraining), confirming that directional decomposition is practically deployable without custom accelerator hardware.

\begin{table}[H]
  \vspace{-2pt}
  \centering
  \scriptsize
  \setlength{\tabcolsep}{3.5pt}
  \caption{\textbf{Measured pre-\(W_O\) projection overhead.}
  Wall-clock overhead relative to unedited baselines on an RTX 6000 Ada (BF16; 2,048 prefill, 512-token decode with CUDA Graph replay; 0.7B training with batch 1, length 2,048).}
  \label{tab:projection-overhead}
  \begingroup
  \renewcommand{\arraystretch}{0.88}
  \begin{tabular}{lccc}
    \toprule
    Workload & Qwen3-0.6B & Qwen3-4B & Qwen3-14B \\
    \midrule
    Prefill & 2.47\% & 1.25\% & 1.11\% \\
    Static graph decode & 0.83\% & 0.36\% & 0.25\% \\
    \midrule
    Training (0.7B) & \multicolumn{3}{c}{3.39\%} \\
    \bottomrule
  \end{tabular}
  \endgroup
  \vspace{-4pt}
\end{table}

\begin{figure*}[!t]
  \vspace{-4pt}
  \centering
  \newsavebox{\headlocalizationfigure}
  \sbox{\headlocalizationfigure}{%
    \includegraphics{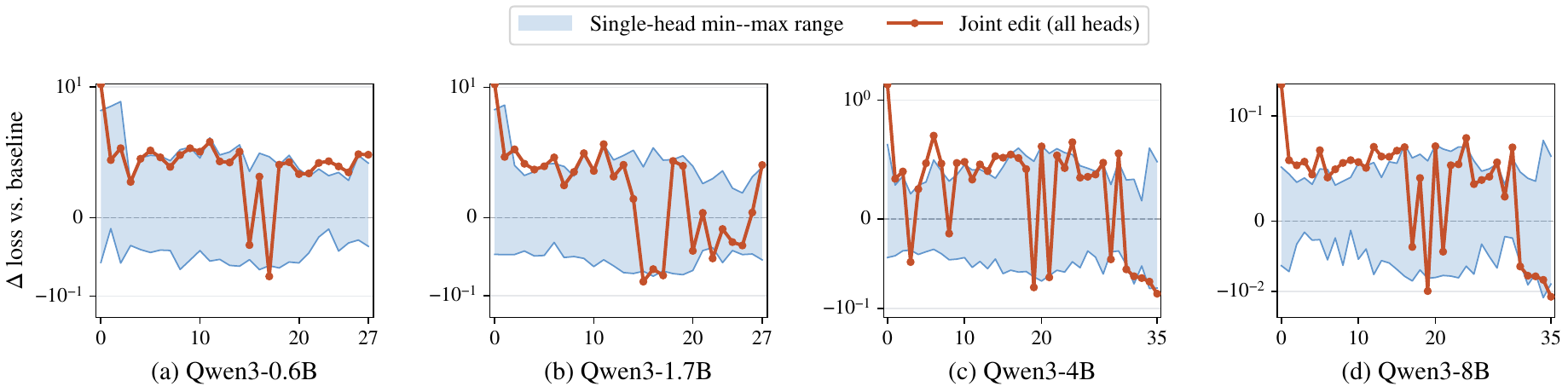}%
  }
  \ifdim\wd\headlocalizationfigure>3\ht\headlocalizationfigure
    \includegraphics[width=0.94\textwidth]{figs/per_head_causal_sensitivity_fullrem/head_delta_loss.pdf}
  \else
    \includegraphics[
      width=0.54\textwidth,
      trim=145bp 358bp 114bp 0bp,
      clip
    ]{figs/per_head_causal_sensitivity_fullrem/head_delta_loss.pdf}

    \vspace{0.1em}
    \makebox[\textwidth][c]{%
      \includegraphics[
        width=0.245\textwidth,
        trim=0bp 172bp 307bp 38bp,
        clip
      ]{figs/per_head_causal_sensitivity_fullrem/head_delta_loss.pdf}%
      \hfill
      \includegraphics[
        width=0.245\textwidth,
        trim=307bp 172bp 0bp 38bp,
        clip
      ]{figs/per_head_causal_sensitivity_fullrem/head_delta_loss.pdf}%
      \hfill
      \includegraphics[
        width=0.245\textwidth,
        trim=0bp 0bp 307bp 218bp,
        clip
      ]{figs/per_head_causal_sensitivity_fullrem/head_delta_loss.pdf}%
      \hfill
      \includegraphics[
        width=0.245\textwidth,
        trim=307bp 0bp 0bp 218bp,
        clip
      ]{figs/per_head_causal_sensitivity_fullrem/head_delta_loss.pdf}%
    }
  \fi
  \vspace{-2pt}
  \caption{\textbf{All-layer head sensitivity across Qwen3 scales.}
  At each layer, orange shows joint full removal from all heads; the blue
  band spans minimum and maximum isolated single-head effects (not a
  confidence interval). $\Delta$ loss is edited minus baseline over 128 fixed C4
  texts (positive is worse, symmetric-log axes).}
  \label{fig:per-head-causal-sensitivity}
  \vspace{-6pt}
\end{figure*}

\newpage
\section{Additional Editing Results}
\label{app:additional-editing}

\paragraph{Zero-shot parallel editing.}
To test whether value-space parallel tolerance is an artifact of Qwen3
or short contexts, Table~\ref{tab:appendix-ruler-full} evaluates retained-scale
editing ($s_{\parallel}=0.5$) under long-context retrieval on RULER-4k across three distinct
model families. Across Gemma-3-12B, Llama-3.2-3B, and Qwen3-4B/8B, scores remain within 0.5
points of matched unedited baselines, confirming that value-space parallel tolerance generalizes
across model families and long-context regimes.

\begin{table}[H]
  \vspace{-5pt}
  \centering
  \scriptsize
  \caption{\textbf{Cross-model retained-scale RULER-4k evaluation.}
  V-Excl.-self preserves direct self contribution and sets
  $s_{\parallel}=0.5$ for the non-self aggregate; \(\Delta\) is edited minus
  baseline (computed before rounding).}
  \label{tab:appendix-ruler-full}
  \setlength{\tabcolsep}{4.5pt}
  \renewcommand{\arraystretch}{0.85}
  \begin{tabular}{lrrr}
    \toprule
    Model & Baseline & V-Excl.-self, \(s_\parallel{=}0.5\) & \(\Delta\) \\
    \midrule
    Gemma-3-1B-IT & 68.0 & 62.8 & -5.3 \\
    Gemma-3-12B-IT & 94.0 & 94.1 & +0.1 \\
    Llama-3.2-3B & 86.9 & 86.7 & -0.2 \\
    Qwen3-0.6B & 79.8 & 75.1 & -4.8 \\
    Qwen3-1.7B & 88.1 & 84.9 & -3.2 \\
    Qwen3-4B & 92.6 & 92.4 & -0.2 \\
    Qwen3-8B & 94.2 & 93.7 & -0.5 \\
    \bottomrule
  \end{tabular}
  \vspace{-5pt}
\end{table}

\paragraph{Head-level localization.}
\label{app:head-localization}
Our primary interventions suppress parallel components across all attention heads jointly.
To determine how head-level sensitivity aggregates, we compare single-head parallel removal against joint all-head removal across four Qwen3 scales (Figure~\ref{fig:per-head-causal-sensitivity}).
This reveals two core properties:
(1)~\textit{Bidirectional single-head sensitivity}: individual head edits produce substantial signed loss shifts (blue envelopes span both degradation and gains), showing distinct, non-negligible directional pulls across heads;
(2)~\textit{Sub-additive joint cancellation}: joint all-head removal (orange curve) does not compound super-linearly, but remains bounded within the single-head min--max envelope and well below the sum of absolute perturbations ($\Delta_{\mathrm{joint}} < \sum\nolimits_h |\Delta_h|$).
Opposing head shifts thus mutually cancel in aggregate.
Only layer~0 exhibits a sharp joint spike, showing initial heads collectively anchor early parallel coordinates.

\section{MLP-Internal Geometry}

\label{app:mlp-internal-carrier}

\begin{table}[!t]
  \centering
  \scriptsize
  \caption{\textbf{Downstream controls for MLP-internal geometry.} Core-MC (9 tasks); MMLU (5-shot); GSM8K-S/F (strict match vs.\ flexible match exact match); TQA (Rouge-L); DROP/CoQA (F1).}
  \label{tab:mlp-internal-downstream}
  \vspace{0.1em}
  
  \textbf{(a) Multiple-choice controls}\\[0.1em]
  \setlength{\tabcolsep}{6pt}
  \renewcommand{\arraystretch}{0.76}
  \begin{tabular}{llrr}
    \toprule
    Model & Setting & Core-MC & MMLU \\
    \midrule
    \multirow{3}{*}{Qwen3-0.6B}
      & Baseline     & 50.94 & 52.40 \\
      & Full No-Para & 50.98 & 52.66 \\
      & Full No-Perp & 36.62 & 26.89 \\
    \midrule
    \multirow{3}{*}{Qwen3-1.7B}
      & Baseline     & 55.19 & 60.21 \\
      & Full No-Para & 55.28 & 60.39 \\
      & Full No-Perp & 37.35 & 24.40 \\
    \bottomrule
  \end{tabular}

  \vspace{0.25em}
  \textbf{(b) Generation and question-answering controls}\\[0.1em]
  \setlength{\tabcolsep}{2.6pt}
  \renewcommand{\arraystretch}{0.76}
  \resizebox{\linewidth}{!}{%
  \begin{tabular}{llrrrrr}
    \toprule
    Model & Setting & GSM8K-S & GSM8K-F & TQA R-L & DROP & CoQA \\
    \midrule
    \multirow{2}{*}{Qwen3-0.6B}
      & Baseline     & 49.43 & 50.42 & 18.98 & 7.76 & 74.24 \\
      & Full No-Para & 49.96 & 50.72 & 15.61 & 7.66 & 74.04 \\
    \midrule
    \multirow{2}{*}{Qwen3-1.7B}
      & Baseline     & 68.76 & 68.92 & 48.34 & 7.32 & 77.06 \\
      & Full No-Para & 68.08 & 68.01 & 48.32 & 7.44 & 77.69 \\
    \bottomrule
  \end{tabular}}
  \vspace{-8pt}
\end{table}

Residual-space editing observes only final MLP output, leaving open whether directional asymmetry precedes down projection.
Inside a gated MLP, carrier $h = \phi(W_{\mathrm{gate}}x)\odot W_{\mathrm{up}}x$ and output $y = \sum_{g=1}^G y_g$ decompose across coordinate groups $I_g = \{(g-1)d+1,\dots,gd\}$.
Each slice $h_g = h[I_g]$ and group contribution $y_g = W_{\mathrm{down}}[:,I_g]h_g$ share dimensionality with residual state $x \in \mathbb{R}^d$.

With directional scaling $\mathcal{P}(u; r) = s_\parallel \frac{u \cdot r}{\|r\|^2}r + s_\perp \left(u - \frac{u \cdot r}{\|r\|^2}r\right)$,
the two internal editing sites in Table~\ref{tab:mlp-internal-carrier} are:
\begin{align}
  \widetilde{y}_{\mathrm{carrier}} &= \sum\nolimits_g W_{\mathrm{down}}[:,I_g]\,\mathcal{P}(h_g; x), \label{eq:mlp-site1} \\[-6pt]
  \widetilde{y}_{\mathrm{contrib}} &= \sum\nolimits_g \mathcal{P}(y_g; h_g). \label{eq:mlp-site2}
\end{align}
Site~1 (\emph{grouped $h$ rel.\ to $x$}) edits intermediate carrier $h_g$ before down projection;
Site~2 (\emph{grouped $y_g$ rel.\ to $h_g$}) edits projected contribution $y_g$ before group summation.
These controls isolate intermediate representation asymmetry from down-projection effects.
On Qwen3-0.6B, parallel removal shifts fixed C4 loss by $\le 0.0301$, while perpendicular removal increases
it by $>13$ points (Table~\ref{tab:mlp-internal-carrier}), confirming directional asymmetry is an intrinsic
property of internal activations.

\begin{table}[H]
  \centering
  \scriptsize
  \caption{\textbf{MLP-internal component removal.}
  C4 $\Delta$loss (lower is better). Full No-Para/No-Perp sets retained scale to zero; no-op matches baseline.}
    \vspace{-8pt}
  \label{tab:mlp-internal-carrier}
  \setlength{\tabcolsep}{5pt}
  \renewcommand{\arraystretch}{0.65}
  \begin{tabular}{lrr}
    \toprule
    Geometry & Full No-Para & Full No-Perp \\
    \midrule
    Residual MLP ($y$ relative to $x$) & +0.1165 & +15.9567 \\
    Grouped $h$ relative to $x$ & +0.0301 & +13.4184 \\
    Grouped $y_g$ relative to $h_g$ & \textbf{+0.0037} & +13.3324 \\
    \bottomrule
  \end{tabular}
  \vspace{-4pt}
\end{table}

\paragraph{Downstream task validation.}
While fixed C4 loss (Table~\ref{tab:mlp-internal-carrier}) confirms internal parallel removal preserves perplexity, cross-entropy averages can mask structured degradation.
Table~\ref{tab:mlp-internal-downstream} audits Full No-Para ($s_{\parallel}=0$) and Full No-Perp ($s_{\perp}=0$) across multiple tasks. 
For GSM8K, \mbox{GSM8K-S} and \mbox{GSM8K-F} denote strict match (requiring exact canonical format) and flexible match (extracting the numerical answer via flexible parsing), respectively.
Across both metrics and all benchmarks, Full No-Para closely tracks unedited baselines, verifying downstream reasoning remains intact without internal parallel components, whereas Full No-Perp collapses catastrophically.
Thus, internal parallel coordinates provide dispensable scaling degrees of freedom, while perpendicular updates encode essential factual and algorithmic knowledge.

\section{Pretraining Setup and Evaluation}
\label{app:training-configs}

\paragraph{Architecture and pretraining setup.}
Pretraining trajectories (Figure~\ref{fig:pretraining}) train autoregressive Transformers from scratch on OpenWebText~\citep{Gokaslan2019OpenWeb} (296M/436M/528M) and FineWeb100BT~\citep{penedo2024the} (1.4B/2.7B) with value-space parallel removal.
Configurations use standard GQA: 1.4B ($d=2048$, 24 layers, 16/4 heads); 2.7B ($d=2560$, 32 layers, 32/8 heads).
Training spans $\approx$104.9B tokens over 200K steps (batch 256, length 2,048; lr $4\times10^{-4}$/$3\times10^{-4}$, 2K warmup, cosine decay, AdamW, 8$\times$H100 in BF16).

\paragraph{Downstream evaluation results.}
Table~\ref{tab:appendix-pretraining-downstream} resolves downstream averages from Table~\ref{tab:pretraining-summary} into benchmark scores across ARC-E, BoolQ, HSwag, OBQA, PIQA, and WinoGr. Value-space removal consistently outperforms the baseline ($+0.7$ on 1.4B, $+1.5$ on 2.7B). Uniform gains confirm that eliminating parallel drift refines capacity without sacrificing task competencies.

\begin{figure}[!t]
  \centering
  \includegraphics[width=0.83\linewidth,keepaspectratio]{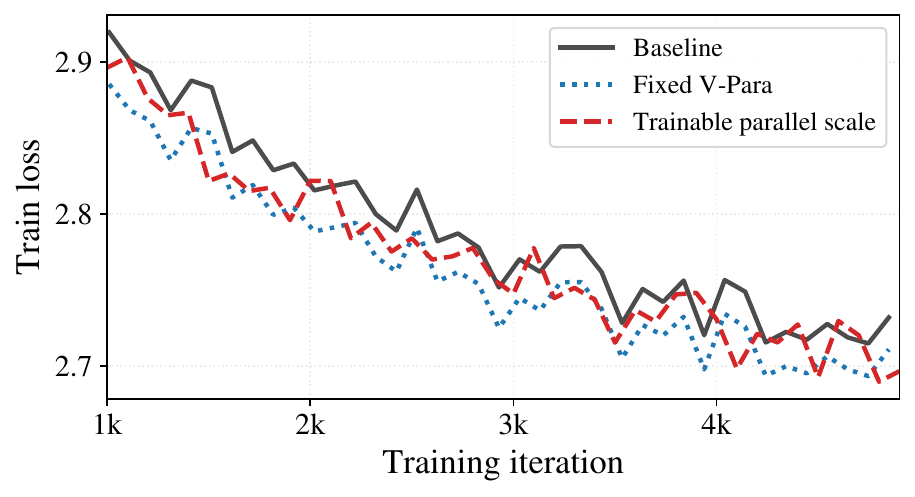}
  \vspace{-4pt}
  \caption{\textbf{Parallel-scale diagnostic on 1.4B pretraining.} Fixed removal ($s_{\parallel}=0$) consistently outperforms learned gating ($s_{\parallel}=\sigma(w^\top x)$).}
  \label{fig:appendix-1p4-trainable-scale}
  \vspace{-5pt}
\end{figure}

\paragraph{Gated parallel-scaling diagnostic.}
To test whether parallel scaling requires learned modulation, we compare fixed removal ($s_{\parallel}=0$) against tokenwise learned gating ($s_{\parallel} = \sigma(w^\top x)$) on 1.4B pretraining (Figure~\ref{fig:appendix-1p4-trainable-scale}). Fixed removal maintains lowest loss throughout, outperforming baseline and learned gating. Dynamic gating converges to higher loss, confirming parallel coordinates introduce unneeded degrees of freedom. Thus, parallel suppression acts as an invariant inductive bias: eliminating parallel drift benefits pretraining from the outset.

\section{Extended Compression Geometry}
\label{app:compression}

While Figure~\ref{fig:local-flip-attn} (\S\ref{sec:compression-error-geometry}) evaluated directional error at attention outputs ($\Delta_{\mathrm{attn}}$), practical compression alters matrices throughout the Transformer block.
Figure~\ref{fig:appendix-compression-breakdown} extends this breakdown to isolated MLP output ($\Delta_{\mathrm{mlp}}$) and combined block update ($\Delta_{\mathrm{block}} = \Delta_{\mathrm{attn}} + \Delta_{\mathrm{mlp}}$).
Across sublayers and the combined update, perpendicular error strictly tracks degradation \mbox{$(\text{AWQ} < \text{unstructured} < 4{:}8 < 2{:}4)$}, while parallel error is non-monotonic.
Table~\ref{tab:local-total-perp-alignment} confirms perpendicular error aligns with total error ($r \ge 0.97$, $\rho \ge 0.90$), accounting for 87.5\%--98.0\% of layer distortion.
Thus, compression distortion is overwhelmingly governed by perpendicular steering; objectives must prioritize preserving directional orientation over scalar norm, explaining why unconstrained pruning damages reasoning.

\begin{table}[!b]
  \centering
  \scriptsize
  \caption{\textbf{Attention-output error alignment.}
  Layerwise correlations on Qwen3-4B.}
  \label{tab:local-total-perp-alignment}
  \setlength{\tabcolsep}{3pt}
  \renewcommand{\arraystretch}{0.74}
  \begin{tabular}{lccccc}
    \toprule
    Setting & \(r(e,e_{\perp})\) & \(\rho(e,e_{\perp})\) &
    \(\rho(e,e_{\parallel})\) & Perp./Total & Para./Total \\
    \midrule
    Quantization & \textbf{0.998} & \textbf{0.990} & 0.596 & \textbf{0.980} & 0.168 \\
    Unstructured & 0.970 & 0.909 & \textbf{0.810} & 0.891 & \textbf{0.439} \\
    4:8 & 0.986 & 0.986 & -0.135 & 0.887 & 0.415 \\
    2:4 & 0.991 & 0.979 & -0.432 & 0.875 & 0.425 \\
    \bottomrule
  \end{tabular}
  \vspace{-4pt}
\end{table}

\paragraph{Alignment with cosine diagnostics.}
Our decomposition directly aligns with cosine diagnostics for layer dropping~\citep{He2026UncoveringTR}.
For update $\Delta = \alpha x + \Delta_\perp$ ($\|\Delta_\perp\|=\beta\|x\|$), cosine similarity is $\cos(x,x+\Delta) \approx 1 - 0.5(\|\Delta_\perp\|/\|x+\Delta_\parallel\|)^2$ since $\|\Delta\| \ll \|x\|$ (Figure~\ref{fig:baseline-update-over-hidden}).
Angular diagnostics thus evaluate perpendicular steering relative to parallel magnitude: layers are safely pruned when perpendicular steering vanishes.

\begin{figure*}[!t]
  \captionsetup{type=table}
  \caption{\textbf{Task-wise downstream scores for pretrained models.}
  Resolving the summary downstream averages reported in Table~\ref{tab:pretraining-summary} (\S\ref{sec:pretraining}) into individual benchmark scores. Avg.\ is unweighted and \(\Delta\)Avg is relative to the baseline at the same model scale. Base,
  Attn, and V-Para denote baseline, residual-space attention parallel removal,
  and full-aggregate value-space parallel removal.}
  \label{tab:appendix-pretraining-downstream}
  \centering
  \small
  \setlength{\tabcolsep}{4pt}
  \begin{tabular}{@{}llrrrrrrrr@{}}
    \toprule
    Model & Setting & ARC-E & BoolQ & HSwag & OBQA & PIQA & WinoGr & Avg & $\Delta$Avg \\
    \midrule
    \multirow{3}{*}{1.4B}
      & Base & 56.0 & \textbf{65.1} & 60.5 & 34.7 & 75.8 & 58.7 & 58.5 & 0.0 \\
      & Attn & 57.1 & 63.9 & 61.3 & 35.3 & 76.0 & 59.1 & 58.8 & +0.3 \\
      & V-Para & \textbf{58.3} & 62.8 & \textbf{62.1} & \textbf{35.9} & \textbf{76.3} & \textbf{59.8} & \textbf{59.2} & \textbf{+0.7} \\
    \midrule
    \multirow{3}{*}{2.7B}
      & Base & 58.4 & 61.2 & 66.0 & 37.1 & 76.5 & 61.8 & 60.2 & 0.0 \\
      & Attn & 59.3 & 62.8 & 66.7 & 37.7 & 77.0 & 62.1 & 60.9 & +0.7 \\
      & V-Para & \textbf{60.4} & \textbf{64.4} & \textbf{67.1} & \textbf{38.2} & \textbf{77.6} & \textbf{62.6} & \textbf{61.7} & \textbf{+1.5} \\
    \bottomrule
  \end{tabular}

  \vspace{1.5em}

  \captionsetup{type=figure}
  \begin{subfigure}[t]{0.485\textwidth}
    \centering
    \includegraphics[width=\linewidth]{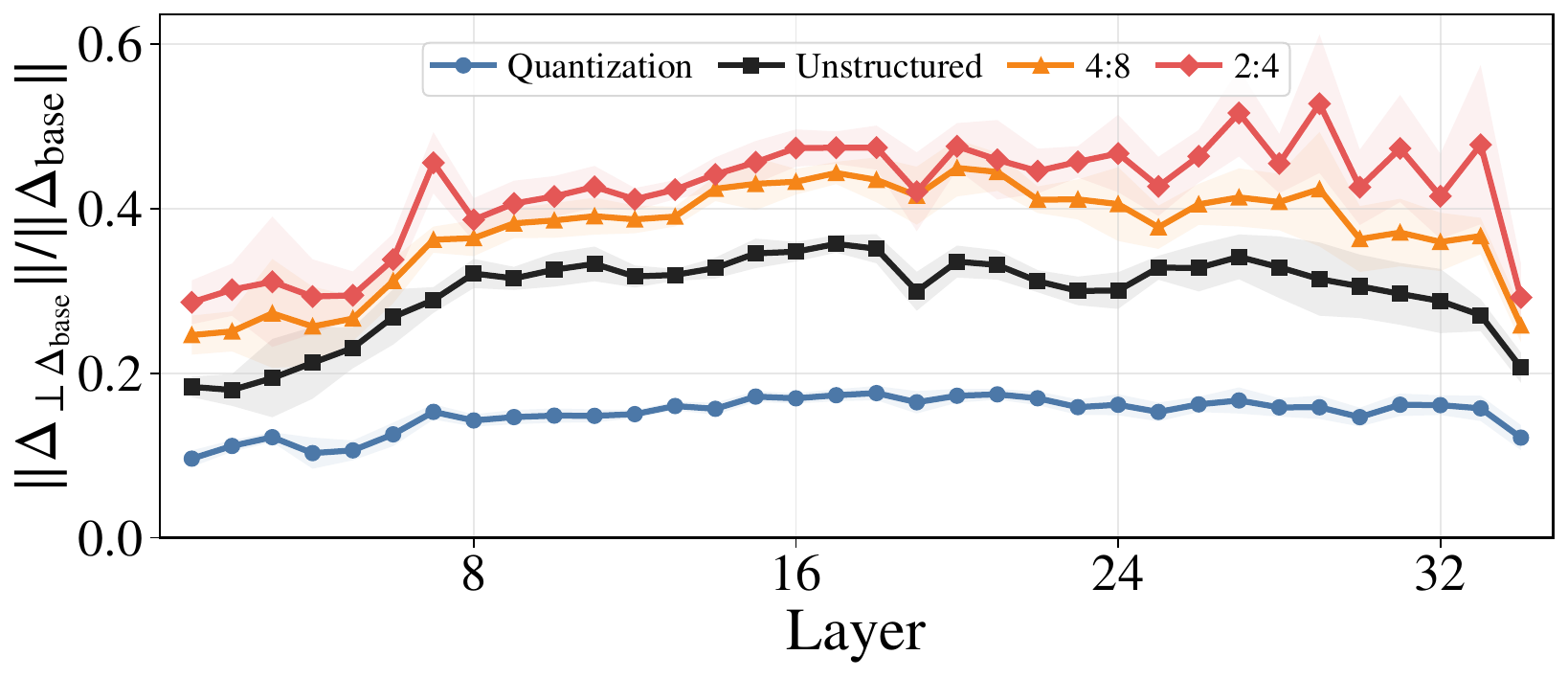}
    \vspace{-0.35em}
    \includegraphics[width=\linewidth]{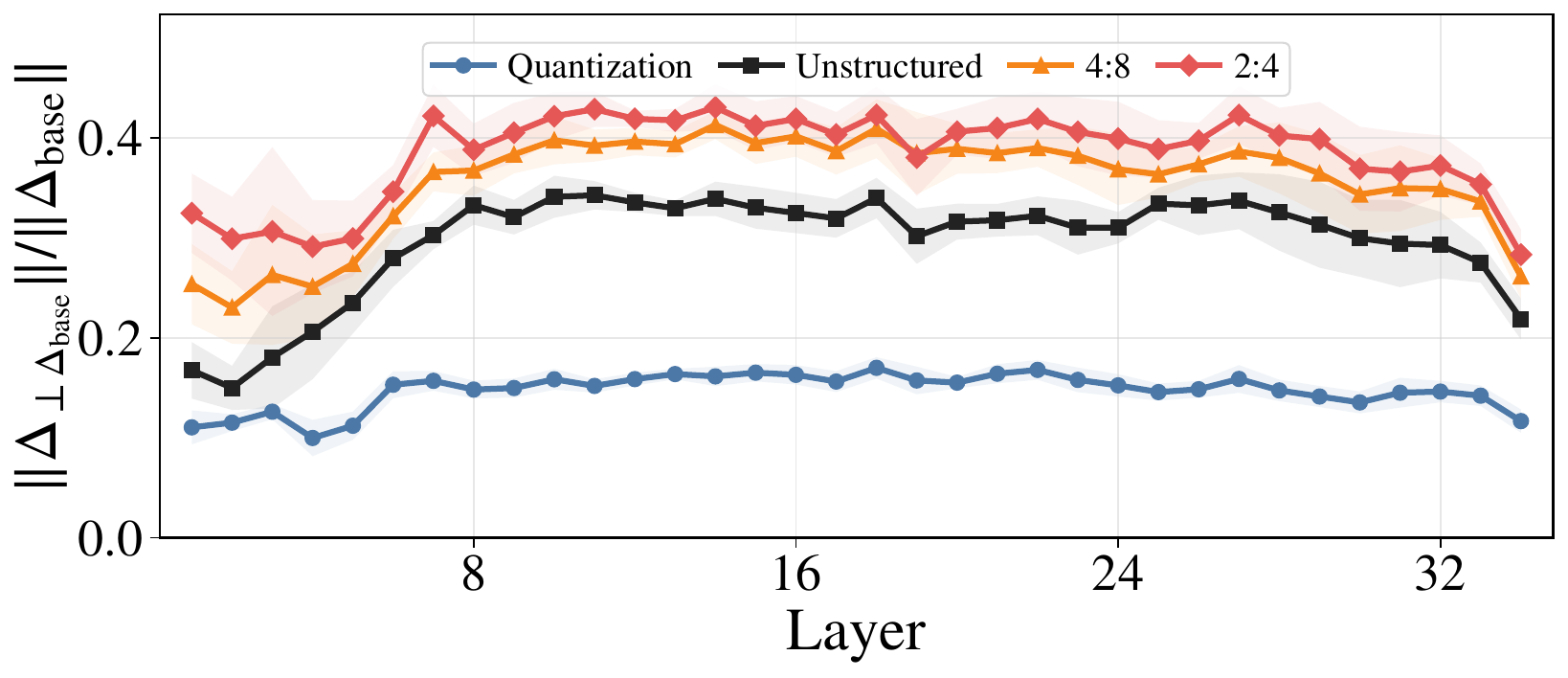}
    \caption{Perpendicular error $\Delta_\perp^{\mathrm{err}}$.}
    \label{fig:appendix-compression-perp}
  \end{subfigure}
  \hfill
  \begin{subfigure}[t]{0.485\textwidth}
    \centering
    \includegraphics[width=\linewidth]{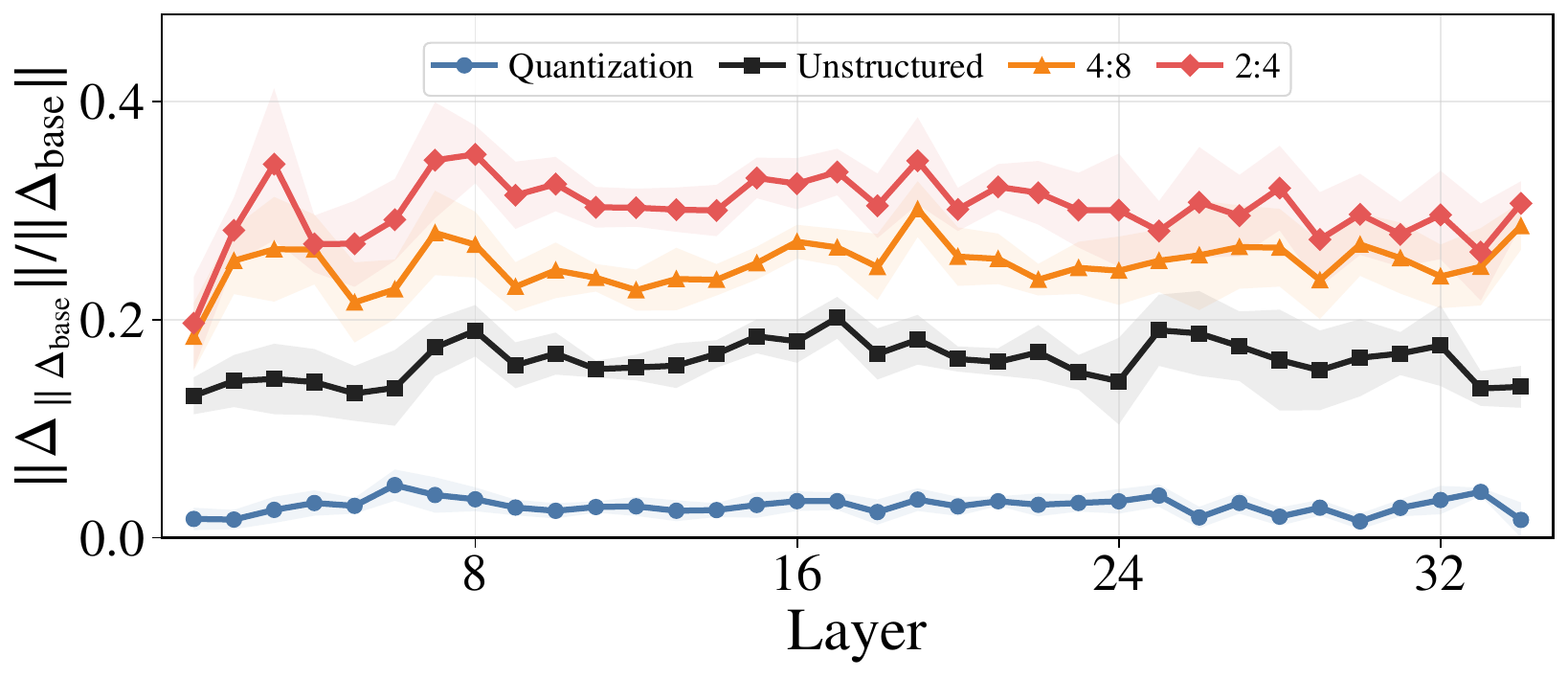}
    \vspace{-0.35em}
    \includegraphics[width=\linewidth]{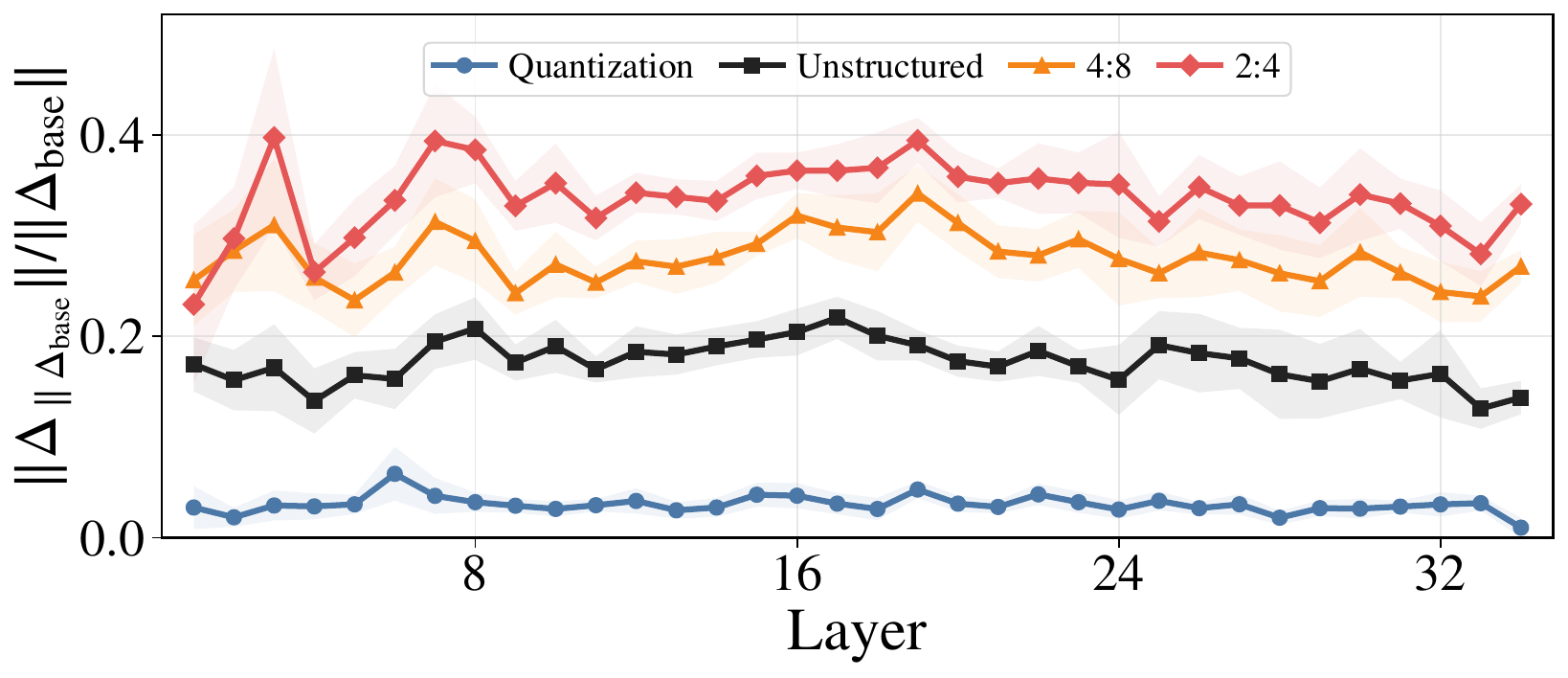}
    \caption{Parallel error $\Delta_\parallel^{\mathrm{err}}$.}
    \label{fig:appendix-compression-para}
  \end{subfigure}
  \caption{\textbf{Compression-geometry breakdown across Block and MLP branches.}
  Evaluation of directional compression distortion across two architectural sites:
  the combined Transformer block update (\(\Delta_{\mathrm{block}} = \Delta_{\mathrm{attn}} + \Delta_{\mathrm{mlp}}\), \textbf{top row})
  and the isolated MLP sub-layer output (\(\Delta_{\mathrm{mlp}}\), \textbf{bottom row}).
  Columns display (a) perpendicular error (\(\Delta_\perp^{\mathrm{err}}\)) and (b) parallel error (\(\Delta_\parallel^{\mathrm{err}}\)),
  respectively, each normalized by the corresponding full baseline-update norm.
  Curves are layerwise means, with bands showing one standard deviation over 12 fixed prompts on
  Qwen3-4B. Settings are 4-bit AWQ and 50\% Wanda pruning with
  unstructured, 4:8, or 2:4 masks.}
  \label{fig:appendix-compression-breakdown}

  \vspace{1.5em}

  \includegraphics[width=0.82\textwidth]{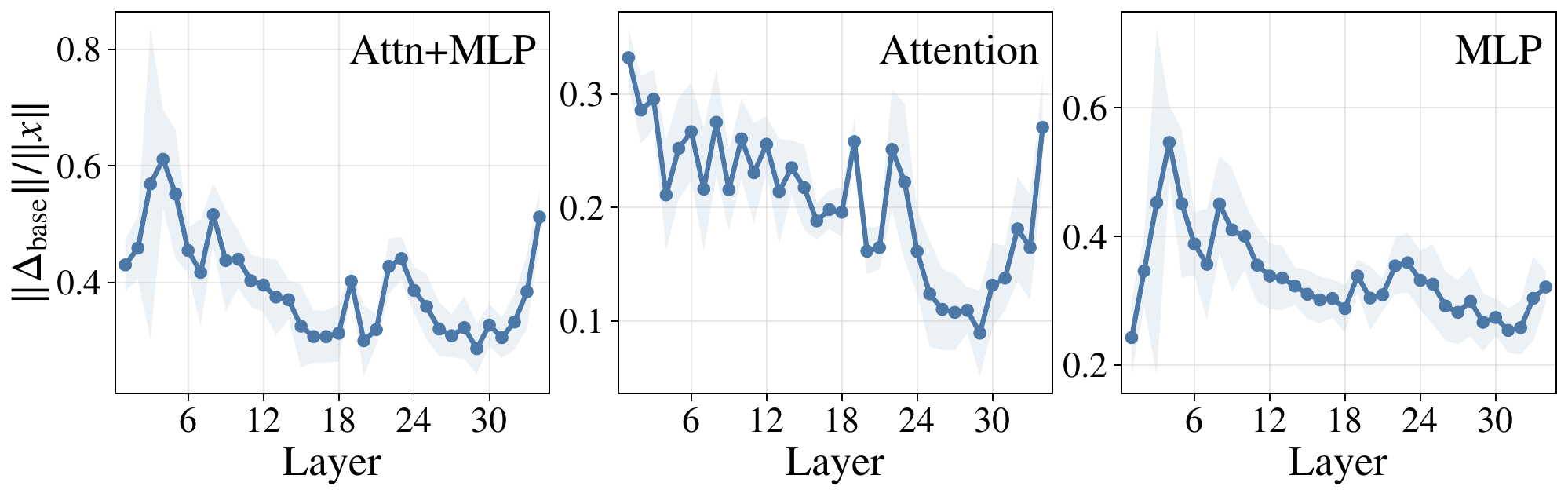}
  \caption{\textbf{Baseline layer updates are small relative to the
  residual stream.} Means and one-standard-deviation bands for
  \(\|\Delta_{\mathrm{base}}\|/\|x\|\) across 12 prompts for Qwen3-4B layers
  1--34, shown for the combined Transformer block update (\(\Delta_{\mathrm{block}} = \Delta_{\mathrm{attn}} + \Delta_{\mathrm{mlp}}\)), attention
  output, and MLP output. Values below one indicate updates smaller than the
  incoming residual stream.}
  \label{fig:baseline-update-over-hidden}
\end{figure*}

\end{document}